\documentclass[]{fairmeta}

\title{\vspace{-1pt} An Empirical Study of Autoregressive Pre-training from Videos}

\author[1,2]{Jathushan Rajasegaran}
\author[2]{Ilija Radosavovic}
\author[2]{Rahul Ravishankar}
\author[1,2]{Yossi Gandelsman}
\author[1]{\\Christoph Feichtenhofer}
\author[1,2]{Jitendra Malik}

\affiliation[1]{Meta FAIR}
\affiliation[2]{UC Berkeley}

\newcommand{\approachName}[1]{\textit{Toto}}

\abstract{
\vspace{-5pt}
We empirically study autoregressive pre-training from videos. To perform our study, we construct a series of autoregressive video models, called \approachName-. We treat videos as sequences of visual tokens and train transformer models to autoregressively predict future tokens. Our models are pre-trained on a diverse dataset of videos and images comprising over 1 trillion visual tokens. We explore different architectural, training, and inference design choices. We evaluate the learned visual representations on a range of downstream tasks including image recognition, video classification, object tracking, and robotics. Our results demonstrate that, despite minimal inductive biases, autoregressive pre-training leads to competitive performance across all benchmarks. Finally, we find that scaling our video models results in similar scaling curves to those seen in language models, albeit with a different rate.
\vspace{-10pt}
}

\metadata[Website]{\url{https://brjathu.github.io/toto}}

\usepackage[T1]{fontenc}    %
\usepackage{url}            %
\usepackage{booktabs}       %
\usepackage{amsfonts}       %
\usepackage{nicefrac}       %
\usepackage{microtype}      %
\usepackage{epsfig, xspace, enumitem}
\usepackage{graphicx, amsmath, amssymb, caption, subcaption, multirow, overpic, textpos, pifont, adjustbox}
\usepackage{xcolor}
\usepackage[utf8x]{inputenc}
\makeatletter
\@namedef{ver@everyshi.sty}{}
\makeatother
\usepackage{pgf, tikz, pgfplots}
\usepackage{makecell}
\usepackage{pgf-pie}
\usepackage{float}
\usepackage{wrapfig}
\usepackage{colortbl}
\usepackage{siunitx}

\definecolor{citecolor}{HTML}{0071BC}
\definecolor{linkcolor}{HTML}{ED1C24}
\definecolor{acceptcolor}{HTML}{74C219}
\definecolor{rejectcolor}{HTML}{DE1616}
\definecolor{qcolor}{HTML}{536872}
\definecolor{demphcolor}{RGB}{100,100,100}
\definecolor{brightlavender}{rgb}{0.75, 0.58, 0.89}
\definecolor{palered}{rgb}{1.00, 0.70, 0.70}
\definecolor{palegreen}{rgb}{0.73, 0.96, 0.67}
\definecolor{paleblue}{rgb}{0.69, 0.84, 1.00}
\definecolor{paleorange}{rgb}{1.00, 0.86, 0.73}
\definecolor{palepurple}{rgb}{0.92, 0.85, 1.00}
\definecolor{paleyellow}{rgb}{1.00, 1.00, 0.50}

\newlength\savewidth

\renewcommand{\paragraph}[1]{\vspace{1.25mm}\noindent\textbf{#1}}
\newcolumntype{L}[1]{>{\raggedright\let\newline\\\arraybackslash\hspace{0pt}}m{#1}}
\newcommand{\app}{\raise.17ex\hbox{$\scriptstyle\sim$}}

\definecolor{lightgray}{rgb}{0.95, 0.95, 0.95}

\definecolor{baselinecolor}{gray}{.9}

\setlist[enumerate]{itemsep=-0.5mm,partopsep=0pt}

\renewcommand{\paragraph}[1]{\vspace{1.25mm}\noindent\textbf{#1}}

\newcolumntype{x}[1]{>{\centering\arraybackslash}p{#1pt}}
\newcolumntype{y}[1]{>{\raggedright\arraybackslash}p{#1pt}}
\newcolumntype{z}[1]{>{\raggedleft\arraybackslash}p{#1pt}}

\begin{document}

\maketitle

\vspace{-0.1cm}
\section{Introduction}
\vspace{-0.1cm}
\label{sec:intro}

In a paper published in 1951, Shannon, having just published the foundational papers of information theory, proposed a ``guessing game'' of {\em next word prediction} to estimate the entropy of English~\citep{shannon1951prediction}. Nearly 70 years later, training a high-capacity transformer network~\citep{vaswani2017attention} on this task, provided the generative pre-training backbone for Large Language Models~\citep{Radford2018, Devlin2019, Radford2019, Brown2020}.

Less well known is the fact that in 1954, Fred Attneave~\citep{attneave1954some} proposed an analog of Shannon’s task for images. To quote “We may divide the picture into arbitrarily small elements which we “transmit” to a subject (S) in a cumulative sequence, having them guess at the color of each successive element until they are correct. 
This method of analysis resembles the scanning process used in television and facsimile systems and accomplishes the like purpose of transforming two spatial dimensions into a single sequence in time”. 

While Attneave was concerned with images, in the context of 2024, we have to note that the “Big Visual Data” is in videos. While there are concerns that most of the text available on the Internet has already been used by the language models, in video we just started on the journey of Big Data exploitation. Despite the successes of autoregressive language and image models, their effectiveness for video modeling remains underexplored.

In this paper, we empirically study autoregressive pre-training from videos. To perform our empirical study, we construct a family of autoregressive video models which we call \approachName~. We treat videos as sequences of visual tokens and train a causal transformer models on next-token prediction task. 
We use causal transformer model with LLaMa~\citep{touvron2023llama} architecture. 
We use  dVAE~\citep{ramesh2021zero} to tokenize frames into discrete tokens. Treating videos as sequences of tokens enables us to jointly train on videos and images using a unified format. We construct a diverse dataset of videos and images comprising over 1 trillion visual tokens. Our models are first pre-trained on this data and then evaluated on downstream tasks. We extract visual representations using attention pooling from relevant layers of the model.

We evaluate our models on various downstream tasks from image and video recognition, video forecasting, semi-supervised tracking, object permanence and robotics tasks in both simulation and real-world. We consider different design choices such as tokenizers including 
 dVAE~\citep{ramesh2021zero}, VQGAN~\citep{rombach2022high} and continuous patch-normalized~\citep{he2022masked} tokens. We also consider different architectures such as LLaMA~\citep{touvron2023llama}, GPT2~\citep{Radford2019} and Mamaba~\cite{gu2023mamba}. Finally we study the compute optimal scaling behaviors of autoregressive video models.

We find that, for tokenization autoregressive models based on discrete and continuous patch-normalized~\citep{he2022masked} tokens perform similarly on ImageNet classification task. 
For efficient pre-training, starting with lower resolution and fine-tuning at higher resolution gives better performance and RoPE~\citep{su2024roformer} helps with adopting to higher resolution. 
For measuring the representation quality in decoder-only models, due to skewed nature of the receptive field we use attention pooling over average pooling.  
We find that in decoder-only models, for all tasks and models sizes the middle layer gives the best performance.
Finally, we study the scaling behaviors of autoregressive vision models, which scales with more compute but still at a slower rate compared to large language models.

\begin{figure*}[t!]
    \centering
    \includegraphics[width=0.95\linewidth]{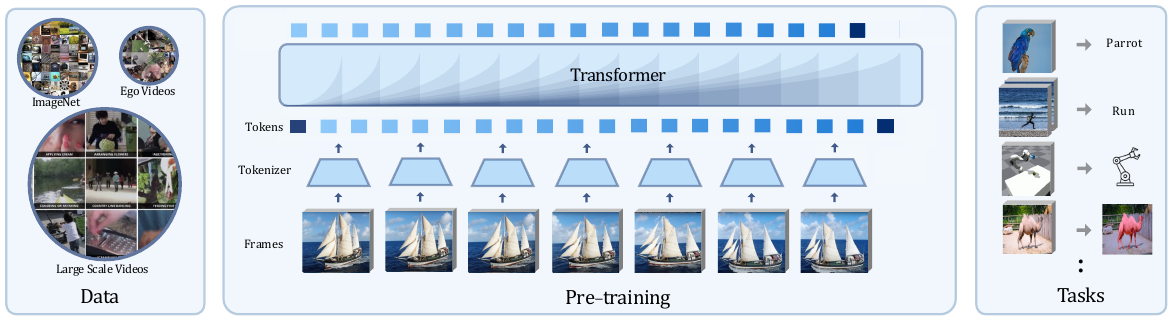}
    \small
    \caption{\textbf{Overall Framework.} Starting with images and video frames from a collection of datasets, we tokenize each frame/image into discrete visual tokens independently.
    We pre-train the transformer by predicting the next visual tokens, with a context length of 4K tokens of images or video frames. Once trained, we take the intermediate representations and evaluate them on various tasks.
    }
    \label{fig:main_fig}
    \vspace{-5mm}
\end{figure*}

\vspace{-0.3cm}
\section{Related work}
\vspace{-0.2cm}
\label{sec:related_work}

\noindent\textbf{Representation Learning for Vision:} 
Over the years self-supervised pre-training has proven to be effective in many areas including language, vision, and robotics. \cite{wu2018unsupervised} and SimCLR~\citep{chen2020simple} showed that instance discrimination training can learn strong discriminative features. MoCo~\citep{he2020momentum} and DINO~\citep{caron2021emerging} showed the effectiveness of strong visual representations on various downstream tasks. Differently, BEiT~\citep{bao2021beit} and MAE~\citep{he2022masked} used masked autoencoding for learning image representations. 
ST-MAE~\citep{feichtenhofer2022masked}and VideoMAE~\citep{wang2023videomae} extended this masked modeling approach to videos, by masking a large amount of tokens during pre-training and predict the masked tokens with a light-weight decoder.

\noindent\textbf{Autoregressive Modeling of Vision:} Generative autoregressive pre-training learns to directly model the data distribution. In language models, generative pre-training has become the standard for training large models. For autoregressive pre-training in vision, rCNN~\citep{ranzato2014video}, PixelCNN~\citep{van2016conditional} and PixelRNN~\citep{van2016pixel} proposed generating pixels one by one using convolution  and bidirectional LSTMs. With the introduction of the transformers~\citep{vaswani2017attention}, ImageTransformers~\citep{parmar2018image} showed generating pixels with causal local attention performs better than previous CNN and RNN-based methods. While all of these methods focused on the generation quality of the pixels, iGPT~\citep{chen2020generative} showed that generative pre-training is also a good way to learn strong visual representations for recognition tasks. \cite{henighan2020scaling} showed scaling behaviors of autoregressive image and video models.
AIM~\citep{el2024scalable} on the other hand uses patch embedding rather than any pre-trained models for tokenization, however, it trains on Data Filtering Networks~\citep{fang2023data} with clip filtered data. 
Compared to these works, we do not use any supervision during our pre-training and utilizes image and videos jointly. VisionMamba~\citep{zhu2024vision} also showed how to utilize sequence models with bidirectional state-space modeling for supervised vision tasks.~\cite{weissenborn2019scaling} showed autoregressive video generation for promotable video generations.

\noindent\textbf{Evaluation of Vision Representations:} Most video pre-training models are evaluated on semantic tasks like ImageNet~\citep{deng2009imagenet} and Kinetics~\citep{kay2017kinetics}. Additionally to the standard evaluation, we evaluate our models on semi-supervised tracking task on DAVIS~\citep{pont20172017}, action forecasting on Ego4D~\citep{grauman2022ego4d}, object permanence on CATER~\citep{girdhar2019cater} and on robot manipulation tasksn simulation~\citep{xiao2022masked} and in the real world~\cite{radosavovic2023robot}.

\newpage
\section{Approach}\label{sec:method}

We train a casual transformer model to predict the next patch tokens in images and videos. 
This is akin to the next token prediction in large language models. From the vast collection of images and videos, every patch is tokenized into a discrete token, and the transformer is trained to predict the next token, using raster scan ordering. We pre-train our models on over one trillion tokens. Finally, we evaluate the learned representations of these models on various downstream tasks including image classification, action classification, action anticipation, video tracking, object permanence, and robotic manipulation tasks. We also study the scaling behaviors of our models for compute optimal training.

\vspace{-0.2cm}
\subsection{Pre-training}

\begin{wrapfigure}{r}{0.34\textwidth}
\vspace{-0.2cm}
\includegraphics[width=0.98\linewidth]{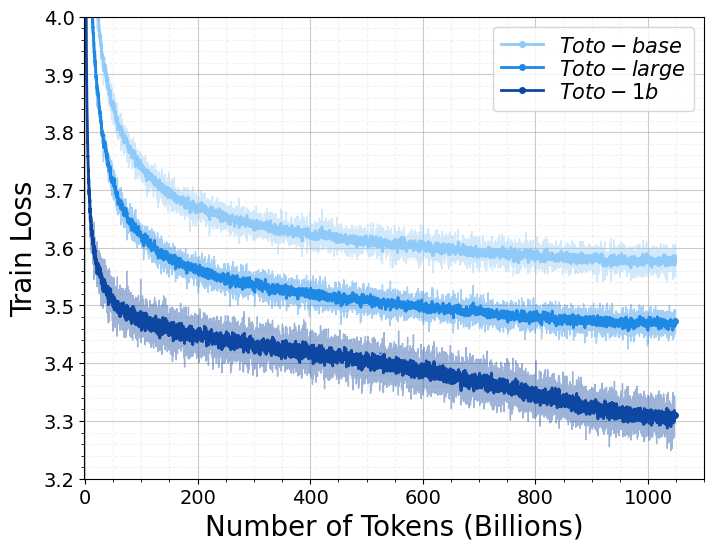}
\caption{\textbf{Training Loss Curves:} We show the training loss curves for base, large, and 1b models trained with tokens from dVAE~\citep{ramesh2021zero} with a vocabulary size of 8k and context length of 4k tokens (equivalent to 16 images or video frames).}
\label{fig:loss_curves_pretrain}
\vspace{-0.5cm}
\end{wrapfigure}

Given a large collection of images and videos, we tokenize all of them into a 1D sequence using raster scan ordering. This produces a dataset of tokens, $\{x^j_1, x^j_2, x^j_3, ..., x^j_n\}$ where $j$ is the sample either from a video or an image and $n$ is the number of tokens in an image or a video. We model the density $p(x)$ as :
\begin{equation}
    p(x^j) = \prod_{i=1}^{n} p(x^j_i | x^j_{i-1}, x^j_{i-2}, ..., x^j_{1}, \theta)
\end{equation}

Here, $\theta$ is the model parameters, which can be optimized by minimizing the negative log-likelihood loss:
\begin{equation}
\label{eq:ntp_loss}
    \mathcal{L}_{\text{pre-train}} = \mathop{\mathbb{E}}_{x^j \sim X} -\log p(x^j).
\end{equation}
Using this loss, we pre-train our models at different sizes on over one visual trillion tokens. These tokens are generated from images and video. Figure~\ref{fig:loss_curves_pretrain} shows the training loss of 3 differently sized models with 120M, 280m and 1.1b parameters.

\vspace{-0.2cm}
\subsection{Architecture}

Our model is a transformer~\citep{vaswani2017attention} with causal attention. We apply recent advancements in language modeling such as pre-norm using RMSNorm~\citep{zhang2019root}, SwiGLU activation~\citep{shazeer2020glu}, and RoPE  positional embeddings~\citep{su2024roformer}, following LLaMa~\citep{touvron2023llama}.

For a model with $L$ layers, we define $H^l$ to be the intermediate representations after layer $l, 0 \leq l \leq L$. The intermediate representations after layer $l+1$, $H^{l+1}$,  defined to be: 
\begin{align}
    \widehat{H}^{l+1} &= {H}^{l} + \texttt{MHSA}(\texttt{RMS-norm}({H}^{l})) \\
    H^{l+1}           &= \widehat{H}^{l+1} + \texttt{MLP}(\texttt{RMS-norm}(\widehat{H}^{l+1})),
\end{align}
Where $\texttt{MHSA}$ is a multi-head self attention layer, $\texttt{MLP}$ is a multi-layer perceptron with SwiGLU activations.

We train our models for the next token prediction task at different scales (base, large and 1b models). For more architecture details see Table~\ref{tbl:models}. We train all these models with a batch size of 1M tokens. We use AdamW~\citep{loshchilov2017decoupled} with a maximum learning rate of $3e{-}4$, and $\beta_1 = 0.9, \beta_2 = 0.95$. 
We decay the learning rate with a cosine schedule, after 2000 warm-up steps~\citep{touvron2023llama}.

\begin{table}[!h]
\begin{center}
\small
\setlength{\tabcolsep}{9pt}
\begin{tabular}{c c c c c c} 
\toprule[0.4mm]
\textbf{Model} & \textbf{Params} & \textbf{Dimension} & \textbf{Heads} & \textbf{Layers} \\ \midrule
base  & 120m & 768  & 12 & 12 \\
large & 280m & 1024 & 16 & 16 \\
1b    & 1.1b & 2048 & 16 & 22 \\
\bottomrule[0.4mm]
\end{tabular}
\end{center}
\vspace{-6pt}
\caption{\textbf{Model Architecture}: We pre-train models at different scales, only on visual tokens from images and videos. 
}
\label{tbl:models}
\vspace{-0.4cm}
\end{table}

\newpage
\subsection{Dataset}

To train our model, we compile a large dataset from a number of different sources. Table~\ref{tbl:datasets} shows the total number of images and videos used for training data, the total number of tokens, as well as the number of hours of videos in each dataset. Together these datasets contain over 100,000 hours of video data and about 2.5 trillion visual tokens. During training, each mini-batch is sampled at different ratios of datasets. Each batch approximately contains 20\% of ImageNet images, 10\% of Ego4D videos, 10\% of Kinetics videos, and 60\% of HowTo100m videos. Our full training utilized about 1 trillion tokens.

\subsection{Tokenization}

We use dVAE tokenizer with a vocabulary of 8k tokens, from Dall-E~\citep{ramesh2021zero} as our tokenizer. Using an image-based tokenizer allows training on both images and videos and testing on respective downstream tasks. While VQGAN~\citep{esser2020taming} tokenizers provide sharper images, these models were trained with perceptual loss~\citep{pmlr-v48-larsen16, johnson2016perceptual}, thus indirectly ingesting ImageNet label information via VGG-net~\citep{simonyan2014very}.

All raw pixel frames or images are tokenized into 256 discrete tokens. 
We take a video and resize it such that its shortest size is $R$ pixels, and then take a random crop of ${R \times R \times T}$, and sample every 4 frames where $T$ is the number of frames. 
We use dVAE~\citep{ramesh2021zero} with the vocabulary of 8k entries to tokenize every frame independently. For dVAE we set $R=128$, to get ${16 \times 16}$ discrete tokens. 
Once every frame is mapped into a set of discrete tokens we have $T \times 256$ tokens per each video. We pre-train all the models with $T=16$, thus all the models are per-trained for a context length of 4096 tokens. 

When training with images and videos, 16 video frames are sampled to create 4k tokens. For images, we randomly sample 16 images and create a sequence of 16 image frames to generate 4k tokens. Finally, we add start and end tokens for each sequence, for videos we use \texttt{[1]} as the start token, and for images we use \texttt{[3]} as the start token, and all sequences have an end token of \texttt{[2]}.

\begin{table}[!t]
\begin{center}
\small
\setlength{\tabcolsep}{15pt}
\renewcommand{\arraystretch}{1.1}
\begin{tabular}{l r r r} 
\toprule[0.4mm]
\textbf{Datasets}    & \textbf{Instances} &  \textbf{Tokens}  & \textbf{Hours} \\ \midrule
ImageNet    & 13.9M & 3.6B  & - \\
Kinetics-600 & 0.53M & 41.3B & 1496  \\
Ego4D       & 52.1K & 103B  & 3750 \\
HowTo100m   & 1.172M & 2560B  & 92627 \\ 
\bottomrule[0.4mm]
\end{tabular}
\end{center}
\caption{\textbf{Pre-training Dataset}: We use both image datasets (Imagenet~\citep{russakovsky2015imagenet}) and video datasets (Kinetics600~\citep{carreira2019short}, Ego4D~\citep{grauman2022ego4d}, HowTo100m~\citep{miech2019howto100m}) with different mixing ratios during the pre-training of our models. The whole training data contains about 100,000 hours of videos. }
\vspace{-0.4cm}
\label{tbl:datasets}
\end{table}

\subsection{Downstream Transfer}\label{sec:method_downstream}

The idea of large pre-trained models is that they were trained at a large compute scale, and then these models can be easily used for various downstream tasks without requiring task-specific design or lots of computing for transfer. The learned representations are general enough to transfer to various tasks. 
We evaluate our models on the intermediate features with linear and attention probing~\citep{pmlr-v97-lee19d}. 

For linear probing the model, we apply global average pooling~\citep{lin2013network} over the tokens from different layers to get the intermediate representation. We train a linear layer on top of this representation on the downstream task. MAE~\citep{he2022masked} or DINO~\citep{caron2021emerging} have a uniform structure when it comes to which token attends to which tokens, however in autoregressive sequence modeling later tokens attent to more tokens than the tokens at the beginning. Due to this skewed nature, equally weighting all the tokens affects the downstream performance. Attention pooling is an alternative to average pooling that allows to dynamically weight the tokens, ideally giving more weight to tokens that see more tokens. This requires learning $W_k$ and $W_v$ matrices and a query token $q$. The query token cross-attends to the intermediate tokens and combines them into a single vector. While this function is not linear anymore, it has been shown to learn better representations in recent works~\citep{el2024scalable}.

\newpage
\section{Experiments}\label{sec:experiments}

We evaluate our pre-trained models on various downstream tasks such as ImageNet classification, Kinetics action recognition, Ego4D action anticipation, Semi-Supervised tracking, and Robotic manipulation tasks. First, we discuss various design choices for pre-training and evaluation strategies for our method. All the models for studying the design choices are \texttt{large} models trained for 400 epochs on the ImageNet-1k dataset.

\subsection{Design Choices}

\noindent\textbf{Tokenizer:} The are various options available for tokenizing an image or a video. We could use discrete tokenizers such as dVAE, and VQGAN, or simple patch-based continuous tokenization. To study the behavior of various tokenizers we pre-train a \approachName X-large model on ImageNet for 400 epochs. Using linear probing at an optimal intermediate layer, we evaluate the accuracy of the models on ImageNet classification task. 

\begin{figure*}[ht!]
    \centering
    \includegraphics[width=1.0\linewidth]{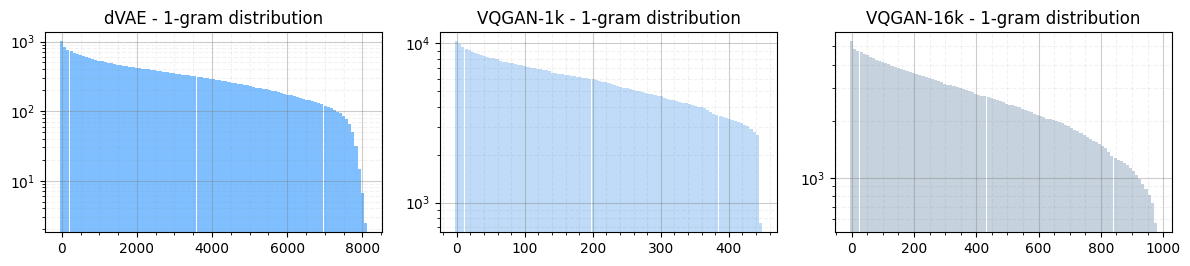}
    \caption{\textbf{1-gram Distribution of Various Tokens:} This Figure shows the distribution of 1-gram tokens of various tokenizers (dVAE~\citep{ramesh2021zero}, VQGAN-1k, VQGAN-16k~\citep{esser2020taming}) on Imagenet validation set. Note that, dVAE has almost full convergence of the tokens while VQGAN has less than 50\% coverage of the tokens.}
    \label{fig:token_1gram}
\end{figure*}

Table~\ref{tab:tokeizers} shows linear probing accuracy when trained with various tokenizers. VQGAN~\citep{esser2020taming} and dVAE~\citep{ramesh2021zero} perform similarly with the same resolutions. However, VQGAN is contaminated with ImageNet label information via perceptual loss. In addition to that, as shown in Figure~\ref{fig:token_1gram}, dVAE tokens have full coverage compared to VQGAN tokens on their 1-gram distributions. Please see the supplementary material for more details. Regressing normalized-patch targets from patch embeddings performs slightly worse than classifying discrete tokens as targets. Additionally, discrete tokens as targets and patch embeddings as inputs perform poorly compared to other methods at the given input-output resolutions. {Overall, Table~\ref{tab:tokeizers} shows that various ways of tokenization have \textit{little effect} on ImageNet linear probing accuracy.}

\begin{table}[!htb]
\centering
\setlength{\tabcolsep}{6pt}
\renewcommand{\arraystretch}{1.2} 
\begin{tabular}{lccc}
\toprule[0.4mm]
\textbf{Input-Target} & \textbf{Tokens} & \textbf{Vocabulary} & \textbf{Top1} \\
\hline
VQGAN-VQGAN            & 16x16 & 16k & 61.3 \\ 
VQGAN-VQGAN            & 16x16 & 1k  & 61.1 \\ 
dVAE-dVAE              & 32x32 & 8k  & 61.2 \\ 
dVAE-dVAE              & 16x16 & 8k  & 53.2 \\ 
patch-patch            & 16x16 & -   & 60.6 \\ 
patch-dVAE             & 16x16 & 8k & 58.5 \\ 
\bottomrule[0.4mm]
\end{tabular}
\vspace{0.2cm}
\caption{\textbf{ImageNet Linear Probing Accuracy with Various Tokenizers:} We compare discrete (dVAE, VQGAN) and patch embedding as input and target for pre-training our models. ImageNet top-1 accuracies are computed by linear probing at the 9th layer of the \texttt{large} model.   }
\label{tab:tokeizers}
\vspace{-0.2cm}
\end{table}

\begin{table}[!htb]
    \sisetup{table-format=1.5}
    \renewcommand\arraystretch{1.2}
\begin{minipage}{.55\linewidth}
  \centering
    \setlength{\tabcolsep}{8pt}
    \begin{tabular}{lcc}
    \toprule[0.4mm]
    \textbf{Method} & \textbf{Compute} & \textbf{Top1} \\
    \hline
    dVAE/16                 & $1.42 \times 10^{17}$   & 53.2 \\ 
    dVAE/32                 & $5.68 \times 10^{17}$   & 61.2 \\ 
    dVAE/16$\rightarrow$32  & $2.13 \times 10^{17}$ & 63.2 \\ 
    dVAE/16$\rightarrow$32$^\dag$ & $2.13 \times 10^{17}$ & 64.4 \\ 
    \bottomrule[0.4mm]
    \end{tabular}
    \vspace{0.2cm}
    \caption{\textbf{Token Resolution:} While the performance is lower for a low-resolution model, when finetuned for next-patch prediction at a higher resolution, its performance surpasses the full-resolution pre-trained model. $^\text{\dag}$ Base values of the RoPE is 50,000.}
    \label{tab:imagenet_res}
\end{minipage}
\hfill  
\begin{minipage}{.4\linewidth}
  \centering
    \setlength{\tabcolsep}{2pt}
    \renewcommand{\arraystretch}{1.8} 
    \begin{tabular}{lccc}
    \toprule[0.4mm]
    \textbf{Method}  & \textbf{Tokens} & \textbf{Pooling} & \textbf{Top1} \\
    \hline 
    dVAE       & 16x16 & Average   & 53.2 \\ 
    dVAE       & 16x16 & Attention & 61.1 \\
    \bottomrule[0.4mm]
    \end{tabular}
    \vspace{0.4cm}
    \caption{\textbf{Attention vs Average Pooling:} When probed at the same layers, attention pooling performs much better than average pooling of intermediate tokens. }
    \label{tab:pooling}
\end{minipage}
\end{table}

\noindent\textbf{How to Probe: } As discussed in Section~\ref{sec:method_downstream} we probe the pre-trained models at the same layer with attention pooling and average pooling, followed by linear layer. Table~\ref{tab:pooling} shows attention pooling performs 7.9\% higher than average pooling on the ImageNet classification task. For attention pooling, we keep the embedding dimension the same as the intermediate feature dimensions.

\noindent\textbf{Resolution: } When training with dVAE tokens, a 256x256 image results in 1024 tokens, this is four times more number of tokens compared to patch embeddings or VQGAN tokens. If we reduce the number of tokens to 256, then the effective image resolution becomes 128x128. Table~\ref{tab:imagenet_res} shows a clear drop in performance when pre-training the model at 128x128 resolution. However, due to the use of relative positional embeddings (RoPE~\citep{su2024roformer}), we can easily finetune the 128x128 (or 16x16 token equivalent) model for higher resolution. Surprisingly, this does better than pre-training at 256x256 resolution and requires only one epoch of finetuning. Not only does this improve the performance, but the pre-training also becomes cheaper compared to full-resolution pre-training. $^\text{\dag}$Additionally, Fine-tuning with higher base values of the RoPE embeddings (50,000) leads to better accuracy.

\noindent\textbf{Architecture: } We train various language models from GPT2~\citep{Radford2019} with absolute sine-cosine positional embeddings, and non-transformer based model Mamba~\citep{gu2023mamba} only using dVAE tokens. We mimicked the GPT2 architecture and do architecture comparisons. We compare these models with LLaMA~\citep{touvron2023llama}. We evaluate linear probing performance at each layer of these models and report the best performance in Table~\ref{tab:models_arch}.

\noindent\textbf{Probing Layer: } When probing the pre-trained models, especially the decoder-only model best performance is observed at the middle layers. This behavior is first observed in iGPT~\citep{chen2020generative}. 
Figure~\ref{fig:probe_layers} shows the peak performance on recognition occurs at about 50\% of the depth of the model. This behavior holds across all model sizes. While in MAE~\citep{he2022masked} and BEiT~\citep{bao2021beit} encoder-decoder models, due to the uneven nature of the encoder and decoder, the best features are observed at the top of the encoder layers. However, on decoder-only models with uniformly distributed layers, the last layers perform worse on recognition tasks, mainly because these layers are trained to reconstruct the input. More probing results with various tokenizers, resolutions, and probing methods are shown in the supplementary material.

\begin{minipage}{0.45\textwidth}
    \small
    \centering
    \begin{tabular}{lcc}
        \toprule[0.4mm]
        \textbf{Model}          & \textbf{Params} & \textbf{Top1}  \\
        \hline
        GPT2~\cite{Radford2019}                & 280 m & 48.5  \\
        Mamba~\cite{gu2023mamba}               & 290 m & 40.7  \\ 
        LLaMA~\cite{touvron2023llama}          & 280 m & 53.2  \\
        \bottomrule[0.4mm]
    \end{tabular}
    \captionof{table}{\textbf{Architecture:} We compare sequence modeling architectures  LLaMA~\cite{touvron2023llama}, GPT2~\cite{Radford2019}, and non-transformer models, Mamba~\cite{gu2023mamba} on ImageNet linear probing task.}
    \label{tab:models_arch}
\end{minipage} \hfill 
\begin{minipage}{0.5\textwidth}
    \centering
    \includegraphics[width=1.0\linewidth]{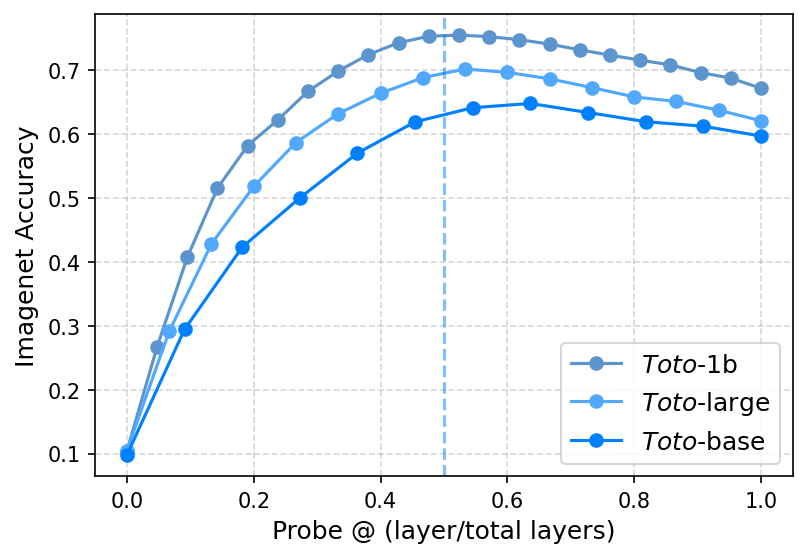}
    \captionof{figure}{\textbf{Probing at Different Layers:} We show the attention-probing performance at each layer of our three models. Peak performance is observed at around 50\% depth of the models.}
    \label{fig:probe_layers}
\end{minipage}

\newpage
\subsection{Image Recognition}

To measure the representation quality of our pre-trained models, we evaluate our models on ImageNet-1k~\citep{deng2009imagenet} classification. We apply a probe at each layer of the model, with attention pooling, and choose the optimal layer with the highest classification accuracy. We fine-tune the pre-trained models further by applying self-supervised next token prediction loss in Eq~\ref{eq:ntp_loss}, together with cross-entropy loss applied for probing layers (with stop-gradients). We train the probing layers for 90 epochs, with a learning rate of $6e^{-5}$. We also use layer decay of 0.9 to reduce the learning rate at the early layers of the model. During this stage, all the models are fine tuned with $32\times32$ token resolution, on the self-supervised loss, and increase the base value of the RoPE~\citep{su2024roformer} embeddings from 10,000 to 50,000 support larger resolution.

Table~\ref{tab:imagenet} shows the ImageNet top-1 accuracy of our \texttt{base, large} and \texttt{1b} models. First, there is a clear difference in terms of classification performance when it comes to discriminative models versus generative models. 
Instance discriminative models such as SimCLR~\citep{chen2020simple}, and DINO~\citep{caron2021emerging} are trained to separate samples from each other and they are designed to perform well on discriminative tasks. On the other hand, generative models are \textit{just} trying to model the data distribution. While achieving comparable performance to other generative models on image recognition, among autoregressive generative models, our model achieved the highest top-1 accuracy. The scaling of data, and the use of tokens instead of pixels, allows our one billion parameter model to achieve similar performance compared to iGPT~\citep{chen2020generative} 7 billion models.

\begin{table}[!htb]
    \centering
    \sisetup{table-format=1.2}
    \renewcommand\arraystretch{1.2}
    \setlength{\tabcolsep}{5pt}
    \begin{tabular}{llcc}
    \toprule[0.4mm]
    \textbf{Method} & \textbf{Arch} & \textbf{\#$\theta$} & \textbf{Top1} \\
    \hline
    \multicolumn{4}{c}{\textcolor{gray}{\textit{Discriminative Approaches}}} \\
    \textcolor{gray}{SimCLR~\citep{chen2020simple}\dag}     & \textcolor{gray}{RN50x2}   & \textcolor{gray}{94}    & \textcolor{gray}{74.2} \\
    \textcolor{gray}{BYOL~\citep{grill2020bootstrap}\dag}   & \textcolor{gray}{RN50x2}   & \textcolor{gray}{94}    & \textcolor{gray}{77.4} \\
    \textcolor{gray}{SwAV~\citep{caron2020unsupervised}\dag}  & \textcolor{gray}{RN50x2}   & \textcolor{gray}{94}    & \textcolor{gray}{73.5} \\
    \textcolor{gray}{DINO~\citep{caron2021emerging}}    & \textcolor{gray}{ViT-B/8}  & \textcolor{gray}{86}    & \textcolor{gray}{80.1} \\
    \textcolor{gray}{DINOv2~\citep{oquab2023dinov2}}    & \textcolor{gray}{ViT-g/14} & \textcolor{gray}{1011}  & \textcolor{gray}{86.4} \\
    \multicolumn{4}{c}{\textit{Generative Approaches}} \\
    BEiT-L~\citep{bao2021beit}        & ViT-L/14 & 307   & 62.2 \\
    AIM~\citep{el2024scalable}        & ViT-1B/14 & 1200   & 80.6 \\ 
    MAE~\citep{he2022masked}          & ViT-H/14 & 632   & 80.9 \\ 
    iGPT-L~\citep{chen2020generative}\dag  & GPT-2    & 1386  & 65.2 \\
    iGPT-XL~\citep{chen2020generative}\dag & GPT-2    & 6801  & 72.0 \\ \midrule
    \approachName X-base             & LLaMA    & 120   & 64.7 \\
    \approachName X-large            & LLaMA    & 280   & 71.1 \\
    \approachName X-1b               & LLaMA    & 1100  & 75.3 \\
    \bottomrule[0.4mm]
    \end{tabular}
    \caption{\textbf{ImageNet Results:} We compare discriminative and generative models on ImageNet~\citep{deng2009imagenet} recognition task. While achieving comparable performance among generative models, our models model achieves the highest accuracy on autoregressive modeling. $^\dag$models are evaluated with linear probing.}
    \label{tab:imagenet}
\end{table}

\subsection{Action Recognition}

We use Kinetics-400 (K400)~\citep{kay2017kinetics} for evaluating our models on action recognition tasks. Similar to ImageNet evaluation, we apply a probe at each layer of the model, with attention pooling, and choose the optimal layer with the highest action classification accuracy.  We also fine-tune the pre-trained models on a self-supervised next-patch prediction task while training the probing layers with a classification loss. All our video models are trained with 16 frames, thus with a context length of 4096 tokens per video. When evaluating videos, we follow the protocol in SlowFast~\citep{feichtenhofer2019slowfast}. Unlike ImageNet where we evaluate the models at 256x256 resolution, on videos we only evaluate our models at 128x128 resolution, to keep the number of tokens in a similar budget. 

Table~\ref{tab:k400} shows the Kinetics-400 top-1 accuracy of our \texttt{base, large} and \texttt{1b} models. Similar to ImageNet results in Table~\ref{tab:imagenet}, we see that discriminately trained models perform better than generative models. Our models achieve comparable performance among generative models, and first to show competitive performance on action recognition with autoregressive generative modeling. All the models are trained and evaluated with 16 frames with a stride of 4 frames.

\begin{table}[!htb]
    \centering
    \renewcommand\arraystretch{1.2}
    \begin{tabular}{llc}
    \toprule[0.4mm]
    \textbf{Method} & \textbf{Arch} & \textbf{Top1}  \\
    \hline
    \multicolumn{3}{c}{\textit{\textcolor{gray}{Discriminative Approaches}}} \\
    \textcolor{gray}{I-JEPA~\citep{assran2023self}}                 & \textcolor{gray}{ViT-H/16}  & \textcolor{gray}{74.5} \\
    \textcolor{gray}{OpenCLIP~\citep{cherti2023reproducible}}       & \textcolor{gray}{ViT-G/14}  & \textcolor{gray}{83.3} \\
    \textcolor{gray}{DINOv2~\citep{oquab2023dinov2}}                & \textcolor{gray}{ViT-g/14}  & \textcolor{gray}{84.4} \\
    \textcolor{gray}{InternVideo~\citep{wang2022internvideo}}       & \textcolor{gray}{-}         & \textcolor{gray}{73.7} \\
    \multicolumn{3}{c}{\textit{Generative Approaches}}  \\
    Hiera~\citep{ryali2023hiera}                  & Hiera-H/14 & 77.0  \\
    MVD~\citep{wang2022masked}                   & ViT-H/14   & 79.4  \\
    VideoMAE~\citep{wang2023videomae}             & ViT-L/14   & 79.8  \\ \hline
    \approachName X-base                         & LLaMA      & 59.3  \\
    \approachName X-large                        & LLaMA      & 65.3  \\
    \approachName X-1b                           & LLaMA      & 74.4  \\
    \bottomrule[0.4mm]
    \end{tabular}
    \caption{\textbf{K400 Results:} We compare discriminative and generative models on Kinetics-400~\citep{kay2017kinetics} action recognition task. While achieving comparable performance among generative models, our models are the first to show the competitive performance on K400 with autoregressive pre-training, and shows scaling with large model sizes. }
    \label{tab:k400}
    \vspace{-0.4cm}
\end{table}

\subsection{Action Forecasting}

While the Kinetics dataset captures internet-style exocentric videos, Ego4D~\citep{grauman2022ego4d} videos capture day-to-day life egocentric videos. A general vision model should be able to reason about both exo and ego-centric videos. Task-wise,  Kinetics requires the model to reason about the action using full context (e.g. the model has seen the action), while the Ego4D short-term action anticipation v1 task requires models to predict future actions from past context. We use our models as the backbone for the pyramid network used in StillFast~\citep{ragusa2023stillfast} extract tokens at 5 layers and fuse them with the pyramid network. We fully fine-tuned our model with self-supervised next-patch loss along with task-related losses, and we observed having self-supervision loss improves overall performance. 
Table~\ref{tab:ego4d} shows the performance of our \texttt{large} model on the Ego4D short-term action anticipation task. This task requires predicting the object to be interacted with (noun) and the type of interaction (verb) as well as time to contact (ttc) from the last seen frame to an estimated time between object-hand contact. As shown in Table~\ref{tab:ego4d}, these tasks are difficult with maximum overall mean-average precision of 2.70.

\begin{table}[!htb]
\centering
\setlength{\tabcolsep}{4pt}
\renewcommand{\arraystretch}{1.05} 
\begin{tabular}{lcccc}
\toprule[0.4mm]
\textbf{Method}          & \textbf{Noun} & \textbf{N+V}  & \textbf{N+TTC} & \textbf{Overall} \\
\hline
FRCNN+Rnd~\citep{grauman2022ego4d}                                         & 17.55 & 1.56 & 3.21  & 0.34    \\
FRCNN+SF~\citep{grauman2022ego4d}                                            & 17.55 & 5.19 & 5.37  & 2.07    \\
Hiera-large~\citep{ryali2023hiera}                      & 14.05 & 6.03 & 4.53  & 2.12    \\
StillFast~\citep{ragusa2023stillfast}              & 16.20 & 7.47 & 4.94  & 2.48    \\
VideoMAE-large~\citep{wang2023videomae}                 & 15.16 & 6.72 & 5.26  & 2.55    \\
MAE-ST-large~\citep{feichtenhofer2022masked}            & 13.71 & 6.63 & 4.94  & 2.60    \\ \midrule
\approachName X-large  & 15.20 & 6.75 & 5.41  & 2.70  \\
\bottomrule[0.4mm]
\end{tabular}
\vspace{0.2cm}
\caption{\textbf{Ego4D Results:} Our model achieves comparable mean-average precision compared to previous work. We compare our method with, FRCNN+Rnd~\citep{grauman2022ego4d}, FRCNN+SF~\citep{grauman2022ego4d}, Hiera~\citep{ryali2023hiera}, StillFast~\citep{ragusa2023stillfast}, VideoMAE~\citep{wang2023videomae}, and MAE-ST~\citep{feichtenhofer2022masked}. }
\label{tab:ego4d}
\end{table}

\begin{figure*}[!htb]
    \centering
    \scalebox{1}[0.8]{
    \includegraphics[width=0.99\linewidth]{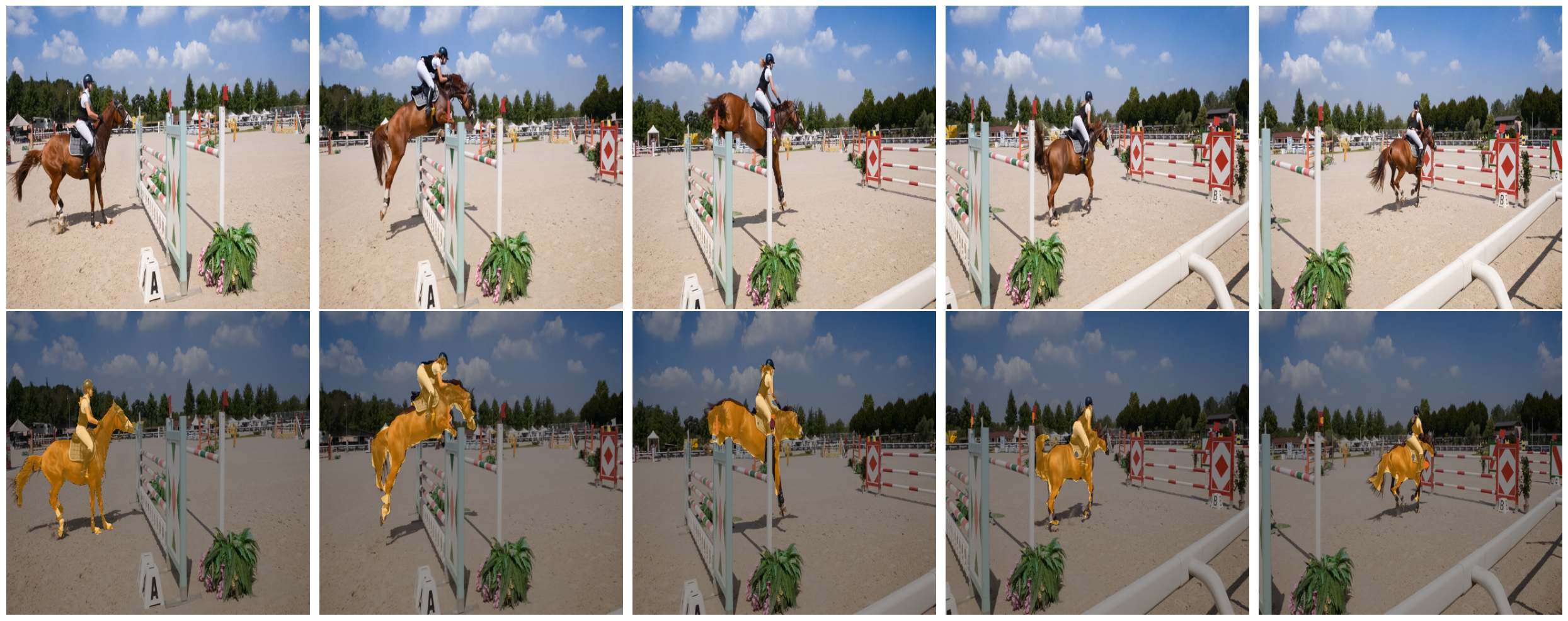}
    }
    \caption{\textbf{Semi-Supervised Tracking:} We follow the protocol in STC~\citep{jabri2020space}, start with the GT segmentation mask, and propagate the labels using the features computed by \approachName X-large. The mask was propagated up to 60 frames without losing much information.}
    \label{fig:tracking}
\end{figure*}

\vspace{0.3cm}
\begin{figure}[!htb]
  \centering
  \begin{subfigure}{0.24\linewidth}
    \includegraphics[width=0.99\textwidth]{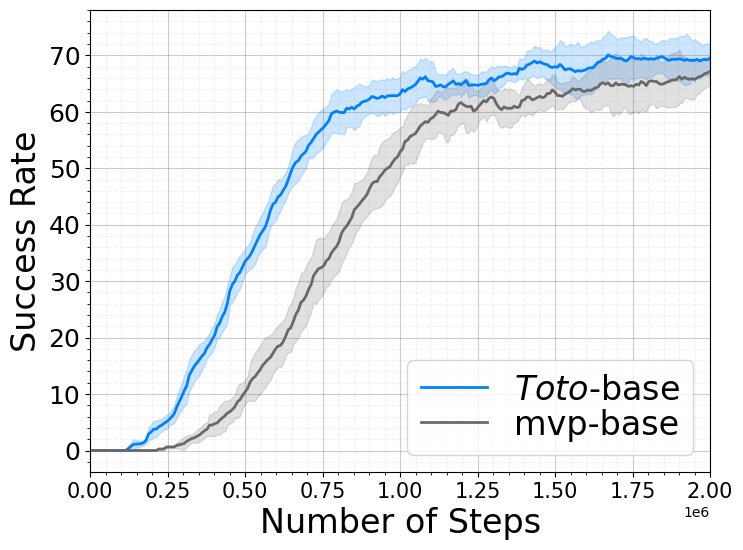}
    \caption{Franka Pick}\label{fig:franka_pick}
  \end{subfigure}
  \hfill
  \begin{subfigure}{0.24\linewidth}
    \includegraphics[width=0.99\textwidth]{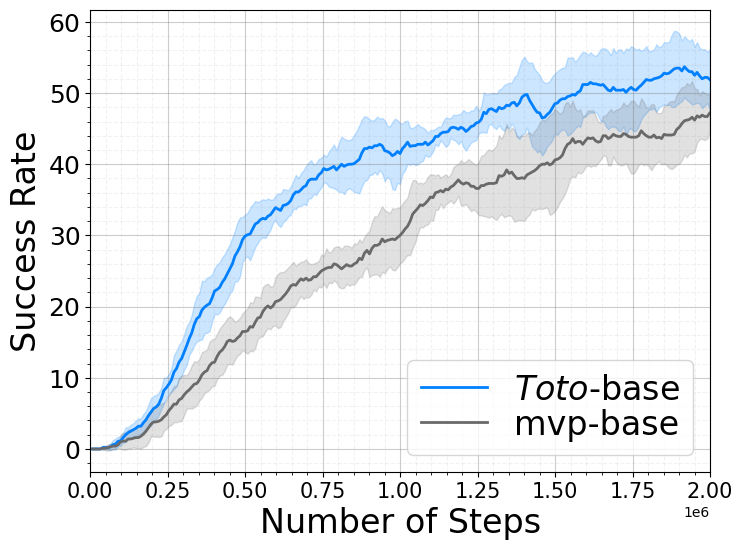}
    \caption{Kuka Pick}\label{fig:kuka_pick}
  \end{subfigure}
  \hfill
  \begin{subfigure}{0.24\linewidth}
    \includegraphics[width=0.99\textwidth]{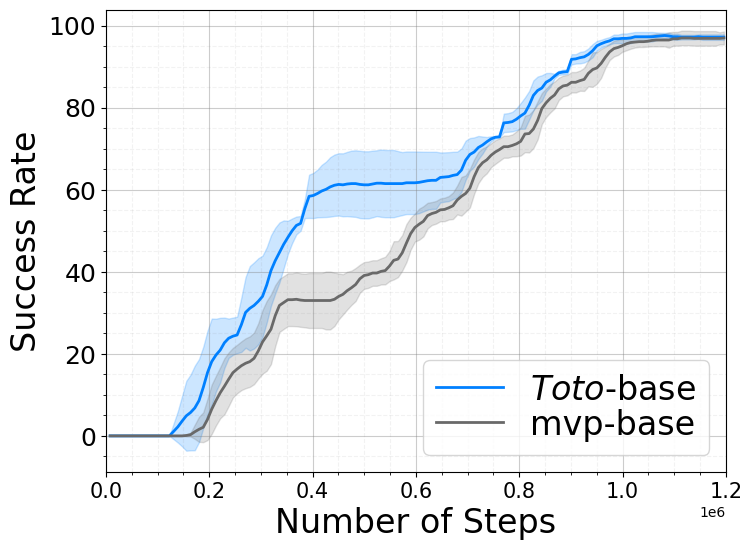}
    \caption{Franka Cabinet}\label{fig:franka_cabinet}
  \end{subfigure}
  \hfill
  \begin{subfigure}{0.24\linewidth}
    \includegraphics[width=0.99\textwidth]{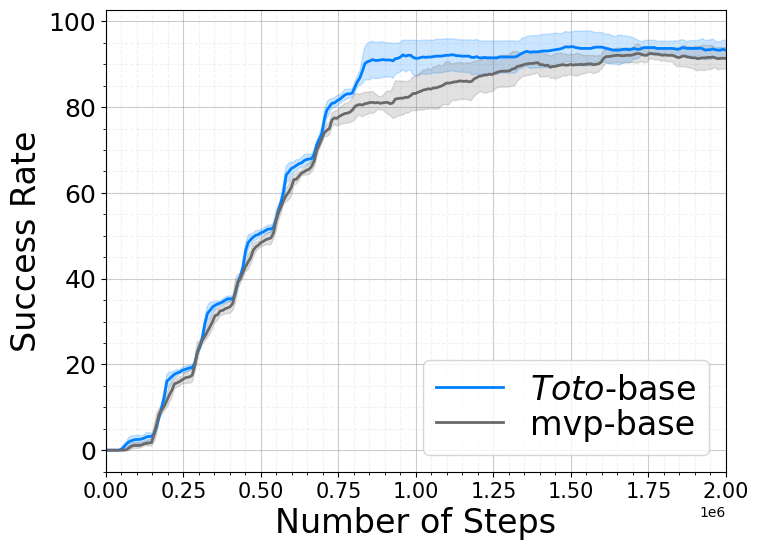}
    \caption{Kuka Cabinet}\label{fig:kuka_cabinet}
  \end{subfigure}
  \hfill
  \caption{\textbf{Robot Manipulation with Reinforcement Learning:} We compare MAE-base~\citep{radosavovic2023real} with \approachName X-base pre-trained models in simulation following~\cite{xiao2022masked}. We evaluate each model the mean success rate over training steps. \approachName X was able to learn these tasks faster than MAE, across two robots and two tasks.}
  \label{fig:robots_all}
\end{figure}

\subsection{Video Tracking}

We study our pre-trained models on label propagation using the protocols in~\citep{jabri2020space} on DAVIS dataset~\citep{pont20172017}. Compared to previous tasks such as classification, and forecasting, this evaluation does not require finetuning or probing of the features. Following \cite{jabri2020space}, we use the features from the last $n$ frames to find the nearest neighbor patch in the current frame, and then propagate the masks from the previous frames to the current frame.Comparison with Dino~\citep{caron2021emerging} and MAE~\citep{he2022masked} is shown in Table~\ref{tbl:tracking_numbers} and qualitative results are shown in Figure~\ref{fig:tracking}.

\begin{table}[!htb]
\small
\centering
\setlength{\tabcolsep}{14pt}
\renewcommand{\arraystretch}{1.2} 
\begin{tabular}{lccc}
\toprule[0.4mm]
\textbf{Method (Res/Patch)}          & \textbf{J\&F} & \textbf{J}  & \textbf{F} \\
\hline
DINO-base (224/8)             & 54.3 & 52.5 & 56.1     \\
DINO-base (224/16)            & 33.1 & 36.2 & 30.1     \\
MAE-base (224/16)            & 31.5 & 34.1 & 28.9     \\ \midrule
\approachName X-base (256/8)                           & 42.0 & 41.2 & 43.1   \\
\approachName X-large (256/8)                         & 44.8 & 44.4 & 45.1   \\
\approachName X-1b   (256/8)                          & 46.1 & 45.8 & 46.4   \\
\approachName X-large (512/8)                    & 62.4 & 59.2 & 65.6   \\
\bottomrule[0.4mm]
\end{tabular}
\captionof{table}{\textbf{DAVIS Tracking:} We report J, F, and {J\&F} scores at the peak layers of each model. We achieves comparable performance as DINO and at large resolution (512), it outperforms all methods.}
\label{tbl:tracking_numbers}
\end{table}

\newpage
\subsection{Robotics}

In this section, we study the effectiveness of our pre-trained representations for robotic manipulation. We consider tasks in both simulation and in the real world. 
Real world experiments needs to run at real time, there for we only use \approachName X-base models, in both setting. Despite being a small model, \approachName X-base can achieve better performance in simulation and on-par performance to state-of-the-art robot models in real world experiments.

\noindent\textbf{Simulation Experiments:}  Following the protocols in MVP~\citep{xiao2022masked}, we use our visual pre-trained models to embed pixel observations. The model is frozen and we only take tokens at an intermediate layer, apply average pooling, and learn the linear layer on top to embed pixel observations. These observations are used to train DAgger policies for 4 different tasks: Franka-pick~\ref{fig:franka_pick}, Kuka-pick~\ref{fig:kuka_pick}, Franka-cabinet~\ref{fig:franka_cabinet}, and Kuka-cabinet tasks~\ref{fig:kuka_cabinet}. Figure~\ref{fig:robots_all} shows the mean success rate over training steps. Compared to the MVP baseline, our model was able to learn these tasks faster with better sample efficiency across robots and tasks. For fair comparisons, we use the best MAE model from MVP~\citep{radosavovic2023real} which is trained on ImageNet~\citep{deng2009imagenet}, Ego4D~\citep{grauman2022ego4d} and 100DOH~\citep{shan2020understanding} datasets.

\begin{wraptable}{r}{7.5cm}
\small
\vspace{-0.4cm}
\setlength{\tabcolsep}{10pt}
\renewcommand{\arraystretch}{1.2}
\centering
\begin{tabular}{lcc}
\toprule[0.4mm]
\textbf{Model}          & \textbf{\# Traj} & \textbf{Success}  \\
\hline
MVP                                    & 240 & 75\%  \\ \hline
\approachName X-base                   & 240 & 63\%  \\
\bottomrule[0.4mm]
\end{tabular}
\caption{\textbf{Robotics, Real-world Experiments:} We compare MVP~\citep{radosavovic2023real} and \approachName X on a Franka cube-picking task in the real world. Features from both models are pre-trained, frozen, and passed into a learning module trained with behavior cloning using the same demonstrations. 
We see that our approach performs comparably to the state-of-the-art vision backbone for robotics, despite not being designed with the robotic application in mind.
}
\vspace{-0.4cm}
\label{tab:real_experiments}
\end{wraptable} 

\begin{figure*}
    \includegraphics[width=0.99\textwidth]{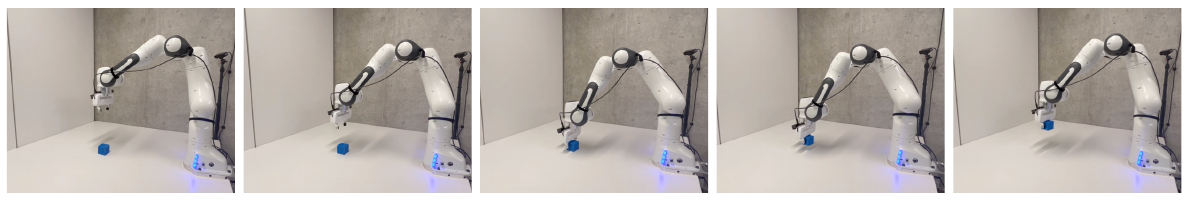}
    \caption{\textbf{Real-world Deployment:} We show an example episode of our policy performing the cube picking task on a Franka robot in the real world. We use \approachName X-base to run the robot at real time, despite being a small model, \approachName X was able to achieve about 63\% success rate in real world setting. }
    \vspace{-0.2cm}
    \label{fig:real_franka_pick}
\end{figure*}

\noindent\textbf{Real-world Experiments:} Next, we evaluate our pre-trained representations in the real world. We follow the setup from~\citep{radosavovic2023real}. We extract vision features using a pre-trained vision encoder and train a controller on top of frozen representations using behavior cloning. Specifically, we consider a cube picking tasks using a 7 DoF Franka robot, shown in Figure~\ref{fig:real_franka_pick}. We use the demonstrations provided by~\citep{radosavovic2023robot}. In Table~\ref{tab:real_experiments} we compare our model to a vision encoder from~\citep{radosavovic2023real}. We report the success rate over 16 trials with variations in object position and orientation. Our model performs favorably to a vision encoder pre-trained for robotics.

\vspace{-0.2cm}
\subsection{Object Permanence}

\begin{wraptable}{r}{7.5cm}
\small
\vspace{-0.4cm}
\centering
\setlength{\tabcolsep}{8.8pt}
\renewcommand{\arraystretch}{1.4} 
\begin{tabular}{lccc}
\toprule[0.4mm]
\textbf{Method}                        &  \bf Model    & \bf 16 & \bf 32 \\ \hline 
V3D                                    & ResNet & 55.2  & 69.7 \\
TFC V3D                                & ResNet & 54.6  & 70.2 \\ \hline 
\approachName X-large                  & LLaMa     & 62.8  & 72.9 \\ 
\bottomrule[0.4mm]
\end{tabular}
\caption{\textbf{Object Permanence:} CATER~\citep{girdhar2019cater} object localization task, where the object is hidden under or obstructed by other  objects. The model is trained to predict its coarse location. Our model performs better than previous methods on snitch localization task at 16, 32 temporal resolutions.} 
\label{tab:cater}
\end{wraptable} 

To quantitatively measure the performance of how well the model understands object permanence, we evaluate our models on CATER localization task~\citep{girdhar2019cater}. Here, a ball is moving in the scene, and the task is to find its location in the 6 by 6 grid. We fine tune our \approachName X-large model on this task at temporal resolutions 16, and 32 frames. In both cases, our pre-trained models were better at localizing the target  compared to models trained specifically for this task, such as V3D~\citep{zhang2022tfcnet}, TFC-V3D~\citep{zhang2022tfcnet}. Table~\ref{tab:cater} shows the performance on the CATER snitch localization task, and \approachName X-large achieve 62.8\% and 70.9\% performance with 16 and 32 frames respectively.

\subsection{Probing Across Layers}

\begin{figure}[!t]
  \centering
  \begin{subfigure}{0.24\linewidth}
    \includegraphics[width=0.99\textwidth]{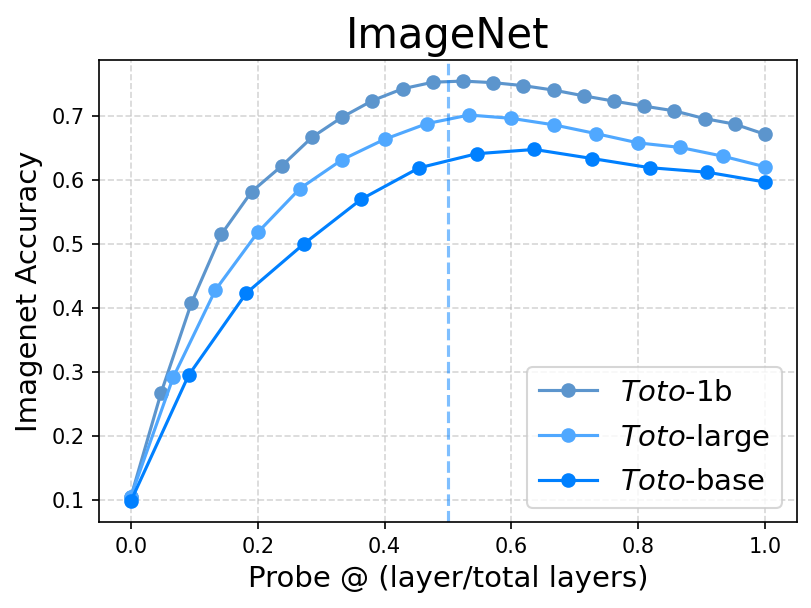}
    \label{fig:imagenet_layers}
  \end{subfigure}
  \hfill
  \begin{subfigure}{0.24\linewidth}
    \includegraphics[width=0.99\textwidth]{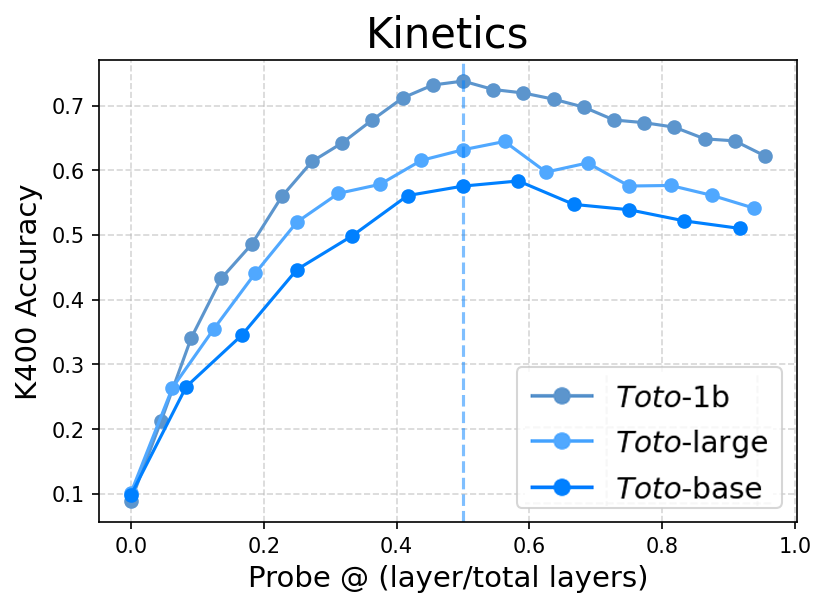}
    \label{fig:k400_layers}
  \end{subfigure}
  \hfill
  \begin{subfigure}{0.24\linewidth}
    \includegraphics[width=0.99\textwidth]{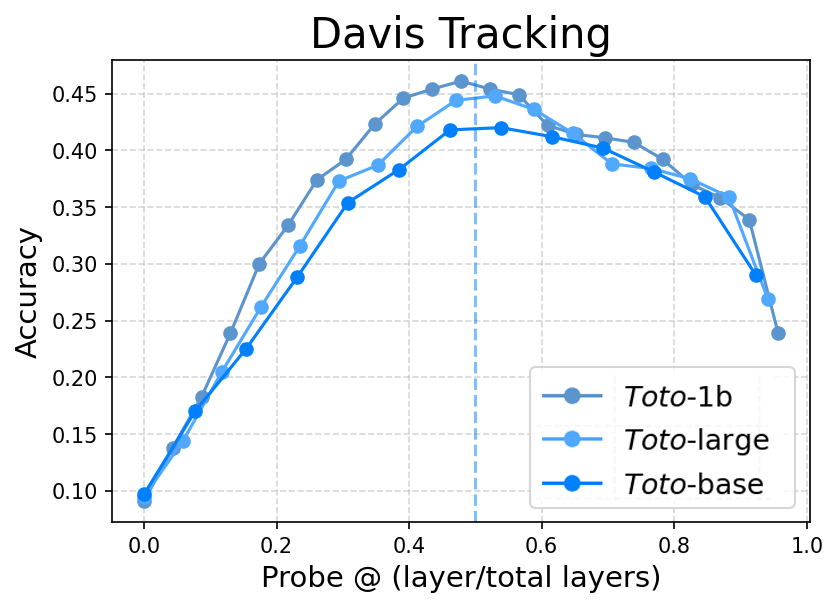}
    \label{fig:davis_layers}
  \end{subfigure}
  \hfill
  \begin{subfigure}{0.24\linewidth}
    \includegraphics[width=0.99\textwidth]{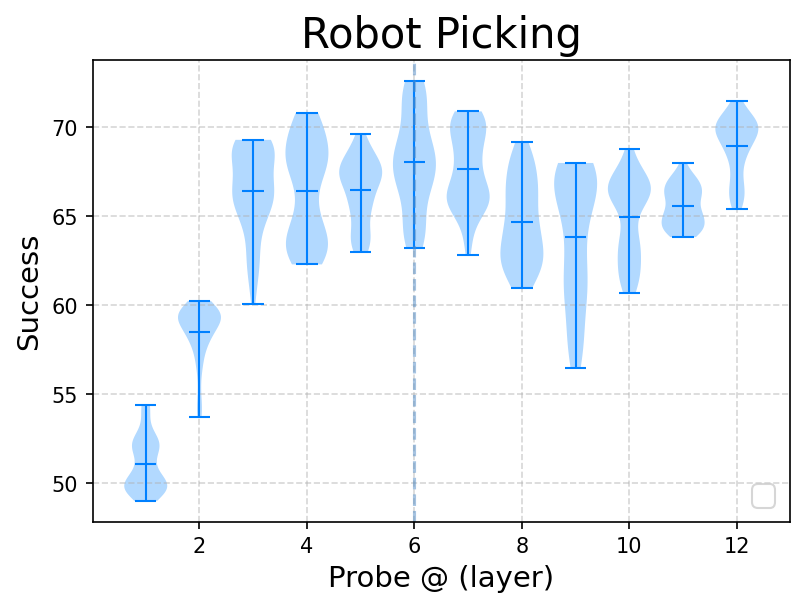}
    \label{fig:franka_layers}
  \end{subfigure}
  \vspace{-0.5cm}
  \caption{\textbf{Probing Across Layers, Models, and Tasks:} We study the behavior of our models across multiple layers and tasks. For image classification, action recognition, and object tracking, all the models behave similarly and peak around 50\% of the model depth. This behavior is observed across all model sizes. Robot tasks show a similar behaviour, where the middle layers perform good at picking the objects, but last layers also perform good as middle layers. These plots suggests, in decoder-only model, first half of the model starts to behave like an encoder, and compress the information, and then rest of the model, projects the compressed semantic features back to input space.}
  \label{fig:probe_all}
\end{figure}

As shown in Figure~\ref{fig:probe_layers} for the ImageNet classification task, different layers of the model contribute to the task differently for the image classification task; this behavior is also observed in iGPT~\citep{chen2020generative}. To study this behavior across multiple tasks, we train probing layers for action recognition, object tracking, and robot manipulation. Figure~\ref{fig:probe_all} shows probing performance across layers, model size, and tasks. It shows that action recognition follows a similar trend to ImageNet classification, having peak performance at the middle of the model stacks. While Object tracking also shares a similar trend with image classification and action recognition, object manipulation shows an interesting trend of the last layers performing well as middle layers from picking objects. Compared to the first three tasks, robot manipulation has a generative nature as a task and can benefit from generative pre-training. In encoder models~\citep{caron2021emerging} or encoder-decoder models~\citep{he2022masked,bao2021beit} the last layer of the encoder has more semantic features. \emph{This may suggest that, in decoder-only model, first half of the model starts to behave like an encoder, and compress the information, and then rest of the model, projects the compressed semantic features back to input space.}

\subsection{Compute Optimal Scaling}

\begin{wrapfigure}{r}{0.50\textwidth}
\vspace{-0.7cm}
\centering
\includegraphics[width=0.99\linewidth,,trim={0 0 0 0}, clip]{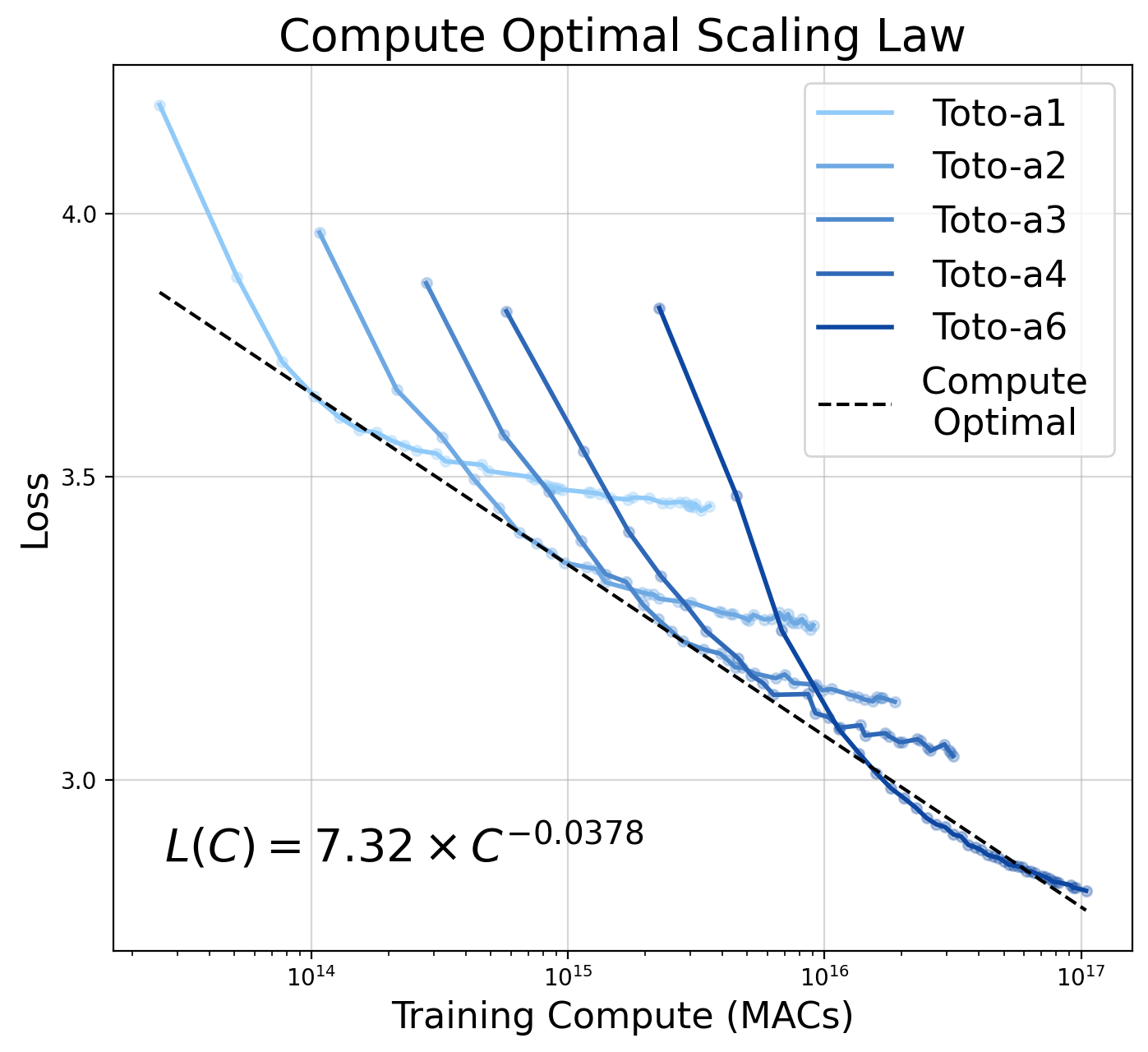}
\small
\vspace{-0.6cm}
\caption{\textbf{Scaling \approachName X:} We train multiple variants of \approachName X, with increasing hidden size and depth, with optimal learning rates. We plot the validation loss vs the compute spent on training in MACs. This shows a clear scaling behavior with optimal compute. }
\label{fig:scale_loss_curves}
\vspace{-0.2cm}
\end{wrapfigure}

\vspace{0.3cm}
We study the scaling behaviors of \approachName X using $\mu$-Parameterization~\citep{yang2022tensorprogramsvtuning}. First we train various models, a1-a6, with linearly increasing hidden size and number of layers (Table~\ref{tbl:a1_a6}). All models use the VQGAN tokenizer~\citep{esser2020taming}. We then optimize the learning rate for these models, with $\mu$-Parameterization~\citep{yang2022tensorprogramsvtuning}. Figure~\ref{fig:mup_loss_curves} shows optimal learning rate of $2^{-7}$ for all of the model widths. Once we find the optimal learning rate, we train a1-a6 models on our data mixture (Table~\ref{tbl:datasets}). Figure~\ref{fig:scale_loss_curves} shows the loss vs training compute of \approachName X models. This shows a clear power law relationship between the compute and validation loss. Based on these experiments \approachName X shows a power law of $L(C) = 7.32 \cdot C^{-0.0378}$. For comparison, the GPT-3 power law relationship~\citep{brown2020language} is $L(C) = 2.57 \cdot C^{-0.048}$. While these are not comparable directly, the scaling coefficients indicate how much change in loss to expect for extra added compute. This suggests that the visual next token prediction models, such as \approachName X, scale but at a slower rate than language models.


\newpage
\section{Limitations}\label{sec:limitations}

Our study suggests several important limitations and opportunities for future work. A significant limitation stems from the use of internet videos, which, unlike carefully curated datasets, introduces challenges related to data quality and diversity. This variance in data quality can impact model performance, especially when compared to models trained on more curated datasets. Another limitation is the use of tokenizer, this makes the learning not end-to-end, and the representation and generation quality is bounded by the quality of the tokenizer, and with quantized vectors, the quality is very much limited, this needs further explorations to build a universal visual tokenizer. Another fundamental limitation is training on videos for next token prediction task. The added redundancy in video frames, can hurt quality of the learned representations. See Appendix~\ref{app:video_training} for more discussion on this topic. Additionally, our exploration of various design choices are based on ImageNet classification. While it does transfer to most of the tasks we considered in this paper, it may not be the optimal configuration for many other tasks. Furthermore, we have not yet fully assessed our method's effectiveness in dealing with dense prediction tasks, fine-grained recognition, or comprehending complex temporal dynamics over extended time frames. These areas represent key opportunities for further research, aiming to broaden the fruitfulness of autoregressive pre-trained models.

\section{Conclusion}\label{sec:discussion}

We empirically studied autoregressive pre-training from images and videos. 
We curated a large video dataset and conducted a large-scale evaluation across a range of diverse tasks, including image recognition, video classification, video forecasting, object tracking, object permanence, and robotic manipulation. 
We performed extensive ablation studies to understand different design choices and compared auto regressive pre-training from videos to strong baselines across different tasks. 
We found that, despite minimal inductive biases, our approach achieves competitive performance across all tasks. Finally, we studied the scaling behavior of visual next token prediction models, and showed it scales with compute, but at a slower rate than text based next token prediction models.

\section{Acknowledgments}\label{sec:acknowledgments}

We thank Andrea Madotto, Po-Yao (Bernie) Huang, and Shiry Ginosar for helpful discussions. We're grateful to Ronghang Hu and Xinlei Chen for their help with TPU setup and code bases. We also thank Baifeng Shi for helping us with robots evaluations. We thank Valentin Gabeur and Neerja Thakkar for their valuable feedback on the paper.

\bibliographystyle{iclr2025_conference}
\bibliography{toto}

\begin{thebibliography}{66}
\providecommand{\natexlab}[1]{#1}
\providecommand{\url}[1]{\texttt{#1}}
\expandafter\ifx\csname urlstyle\endcsname\relax
  \providecommand{\doi}[1]{doi: #1}\else
  \providecommand{\doi}{doi: \begingroup \urlstyle{rm}\Url}\fi

\bibitem[Assran et~al.(2023)Assran, Duval, Misra, Bojanowski, Vincent, Rabbat, LeCun, and Ballas]{assran2023self}
Mahmoud Assran, Quentin Duval, Ishan Misra, Piotr Bojanowski, Pascal Vincent, Michael Rabbat, Yann LeCun, and Nicolas Ballas.
\newblock Self-supervised learning from images with a joint-embedding predictive architecture.
\newblock In \emph{Proceedings of the IEEE/CVF Conference on Computer Vision and Pattern Recognition}, pp.\  15619--15629, 2023.

\bibitem[Attneave(1954)]{attneave1954some}
Fred Attneave.
\newblock Some informational aspects of visual perception.
\newblock \emph{Psychological review}, 1954.

\bibitem[Bao et~al.(2021)Bao, Dong, Piao, and Wei]{bao2021beit}
Hangbo Bao, Li~Dong, Songhao Piao, and Furu Wei.
\newblock Beit: Bert pre-training of image transformers.
\newblock \emph{arXiv preprint arXiv:2106.08254}, 2021.

\bibitem[Brown(2020)]{brown2020language}
Tom~B Brown.
\newblock Language models are few-shot learners.
\newblock \emph{arXiv preprint arXiv:2005.14165}, 2020.

\bibitem[Brown et~al.(2020)Brown, Mann, Ryder, Subbiah, Kaplan, Dhariwal, Neelakantan, Shyam, Sastry, Askell, et~al.]{Brown2020}
Tom~B Brown, Benjamin Mann, Nick Ryder, Melanie Subbiah, Jared Kaplan, Prafulla Dhariwal, Arvind Neelakantan, Pranav Shyam, Girish Sastry, Amanda Askell, et~al.
\newblock Language models are few-shot learners.
\newblock \emph{NeurIPS}, 2020.

\bibitem[Caron et~al.(2020)Caron, Misra, Mairal, Goyal, Bojanowski, and Joulin]{caron2020unsupervised}
Mathilde Caron, Ishan Misra, Julien Mairal, Priya Goyal, Piotr Bojanowski, and Armand Joulin.
\newblock Unsupervised learning of visual features by contrasting cluster assignments.
\newblock \emph{Advances in neural information processing systems}, 33:\penalty0 9912--9924, 2020.

\bibitem[Caron et~al.(2021)Caron, Touvron, Misra, J{\'e}gou, Mairal, Bojanowski, and Joulin]{caron2021emerging}
Mathilde Caron, Hugo Touvron, Ishan Misra, Herv{\'e} J{\'e}gou, Julien Mairal, Piotr Bojanowski, and Armand Joulin.
\newblock Emerging properties in self-supervised vision transformers.
\newblock In \emph{Proceedings of the IEEE/CVF international conference on computer vision}, pp.\  9650--9660, 2021.

\bibitem[Carreira et~al.(2019)Carreira, Noland, Hillier, and Zisserman]{carreira2019short}
Joao Carreira, Eric Noland, Chloe Hillier, and Andrew Zisserman.
\newblock A short note on the kinetics-700 human action dataset.
\newblock \emph{arXiv preprint arXiv:1907.06987}, 2019.

\bibitem[Chen et~al.(2020{\natexlab{a}})Chen, Radford, Child, Wu, Jun, Luan, and Sutskever]{chen2020generative}
Mark Chen, Alec Radford, Rewon Child, Jeffrey Wu, Heewoo Jun, David Luan, and Ilya Sutskever.
\newblock Generative pretraining from pixels.
\newblock In \emph{International conference on machine learning}, pp.\  1691--1703. PMLR, 2020{\natexlab{a}}.

\bibitem[Chen et~al.(2020{\natexlab{b}})Chen, Kornblith, Norouzi, and Hinton]{chen2020simple}
Ting Chen, Simon Kornblith, Mohammad Norouzi, and Geoffrey Hinton.
\newblock A simple framework for contrastive learning of visual representations.
\newblock In \emph{International conference on machine learning}, pp.\  1597--1607. PMLR, 2020{\natexlab{b}}.

\bibitem[Cherti et~al.(2023)Cherti, Beaumont, Wightman, Wortsman, Ilharco, Gordon, Schuhmann, Schmidt, and Jitsev]{cherti2023reproducible}
Mehdi Cherti, Romain Beaumont, Ross Wightman, Mitchell Wortsman, Gabriel Ilharco, Cade Gordon, Christoph Schuhmann, Ludwig Schmidt, and Jenia Jitsev.
\newblock Reproducible scaling laws for contrastive language-image learning.
\newblock In \emph{Proceedings of the IEEE/CVF Conference on Computer Vision and Pattern Recognition}, pp.\  2818--2829, 2023.

\bibitem[Deng et~al.(2009)Deng, Dong, Socher, Li, Li, and Fei-Fei]{deng2009imagenet}
Jia Deng, Wei Dong, Richard Socher, Li-Jia Li, Kai Li, and Li~Fei-Fei.
\newblock Imagenet: A large-scale hierarchical image database.
\newblock In \emph{2009 IEEE conference on computer vision and pattern recognition}, pp.\  248--255. Ieee, 2009.

\bibitem[Devlin et~al.(2019)Devlin, Chang, Lee, and Toutanova]{Devlin2019}
Jacob Devlin, Ming-Wei Chang, Kenton Lee, and Kristina Toutanova.
\newblock Bert: Pre-training of deep bidirectional transformers for language understanding.
\newblock 2019.

\bibitem[El-Nouby et~al.(2024)El-Nouby, Klein, Zhai, Bautista, Toshev, Shankar, Susskind, and Joulin]{el2024scalable}
Alaaeldin El-Nouby, Michal Klein, Shuangfei Zhai, Miguel~Angel Bautista, Alexander Toshev, Vaishaal Shankar, Joshua~M Susskind, and Armand Joulin.
\newblock Scalable pre-training of large autoregressive image models.
\newblock \emph{arXiv preprint arXiv:2401.08541}, 2024.

\bibitem[Esser et~al.(2020)Esser, Rombach, and Ommer]{esser2020taming}
Patrick Esser, Robin Rombach, and Bj{\"o}rn Ommer.
\newblock Taming transformers for high-resolution image synthesis. 2021 ieee.
\newblock In \emph{CVF Conference on Computer Vision and Pattern Recognition (CVPR)}, pp.\  12868--12878, 2020.

\bibitem[Fang et~al.(2023)Fang, Jose, Jain, Schmidt, Toshev, and Shankar]{fang2023data}
Alex Fang, Albin~Madappally Jose, Amit Jain, Ludwig Schmidt, Alexander Toshev, and Vaishaal Shankar.
\newblock Data filtering networks.
\newblock \emph{arXiv preprint arXiv:2309.17425}, 2023.

\bibitem[Feichtenhofer et~al.(2019)Feichtenhofer, Fan, Malik, and He]{feichtenhofer2019slowfast}
Christoph Feichtenhofer, Haoqi Fan, Jitendra Malik, and Kaiming He.
\newblock Slowfast networks for video recognition.
\newblock In \emph{Proceedings of the IEEE/CVF international conference on computer vision}, pp.\  6202--6211, 2019.

\bibitem[Feichtenhofer et~al.(2022)Feichtenhofer, Li, He, et~al.]{feichtenhofer2022masked}
Christoph Feichtenhofer, Yanghao Li, Kaiming He, et~al.
\newblock Masked autoencoders as spatiotemporal learners.
\newblock \emph{Advances in neural information processing systems}, 35:\penalty0 35946--35958, 2022.

\bibitem[Girdhar \& Ramanan(2019)Girdhar and Ramanan]{girdhar2019cater}
Rohit Girdhar and Deva Ramanan.
\newblock Cater: A diagnostic dataset for compositional actions and temporal reasoning.
\newblock \emph{arXiv preprint arXiv:1910.04744}, 2019.

\bibitem[Grauman et~al.(2022)Grauman, Westbury, Byrne, Chavis, Furnari, Girdhar, Hamburger, Jiang, Liu, Liu, et~al.]{grauman2022ego4d}
Kristen Grauman, Andrew Westbury, Eugene Byrne, Zachary Chavis, Antonino Furnari, Rohit Girdhar, Jackson Hamburger, Hao Jiang, Miao Liu, Xingyu Liu, et~al.
\newblock Ego4d: Around the world in 3,000 hours of egocentric video.
\newblock In \emph{Proceedings of the IEEE/CVF Conference on Computer Vision and Pattern Recognition}, pp.\  18995--19012, 2022.

\bibitem[Grill et~al.(2020)Grill, Strub, Altch{\'e}, Tallec, Richemond, Buchatskaya, Doersch, Avila~Pires, Guo, Gheshlaghi~Azar, et~al.]{grill2020bootstrap}
Jean-Bastien Grill, Florian Strub, Florent Altch{\'e}, Corentin Tallec, Pierre Richemond, Elena Buchatskaya, Carl Doersch, Bernardo Avila~Pires, Zhaohan Guo, Mohammad Gheshlaghi~Azar, et~al.
\newblock Bootstrap your own latent-a new approach to self-supervised learning.
\newblock \emph{Advances in neural information processing systems}, 33:\penalty0 21271--21284, 2020.

\bibitem[Gu \& Dao(2023)Gu and Dao]{gu2023mamba}
Albert Gu and Tri Dao.
\newblock Mamba: Linear-time sequence modeling with selective state spaces.
\newblock \emph{arXiv preprint arXiv:2312.00752}, 2023.

\bibitem[He et~al.(2020)He, Fan, Wu, Xie, and Girshick]{he2020momentum}
Kaiming He, Haoqi Fan, Yuxin Wu, Saining Xie, and Ross Girshick.
\newblock Momentum contrast for unsupervised visual representation learning.
\newblock In \emph{Proceedings of the IEEE/CVF conference on computer vision and pattern recognition}, pp.\  9729--9738, 2020.

\bibitem[He et~al.(2022)He, Chen, Xie, Li, Doll{\'a}r, and Girshick]{he2022masked}
Kaiming He, Xinlei Chen, Saining Xie, Yanghao Li, Piotr Doll{\'a}r, and Ross Girshick.
\newblock Masked autoencoders are scalable vision learners.
\newblock In \emph{Proceedings of the IEEE/CVF conference on computer vision and pattern recognition}, pp.\  16000--16009, 2022.

\bibitem[Henighan et~al.(2020)Henighan, Kaplan, Katz, Chen, Hesse, Jackson, Jun, Brown, Dhariwal, Gray, et~al.]{henighan2020scaling}
Tom Henighan, Jared Kaplan, Mor Katz, Mark Chen, Christopher Hesse, Jacob Jackson, Heewoo Jun, Tom~B Brown, Prafulla Dhariwal, Scott Gray, et~al.
\newblock Scaling laws for autoregressive generative modeling.
\newblock \emph{arXiv preprint arXiv:2010.14701}, 2020.

\bibitem[Jabri et~al.(2020)Jabri, Owens, and Efros]{jabri2020space}
Allan Jabri, Andrew Owens, and Alexei Efros.
\newblock Space-time correspondence as a contrastive random walk.
\newblock \emph{Advances in neural information processing systems}, 33:\penalty0 19545--19560, 2020.

\bibitem[Johnson et~al.(2016)Johnson, Alahi, and Fei-Fei]{johnson2016perceptual}
Justin Johnson, Alexandre Alahi, and Li~Fei-Fei.
\newblock Perceptual losses for real-time style transfer and super-resolution.
\newblock In \emph{Computer Vision--ECCV 2016: 14th European Conference, Amsterdam, The Netherlands, October 11-14, 2016, Proceedings, Part II 14}, pp.\  694--711. Springer, 2016.

\bibitem[Kay et~al.(2017)Kay, Carreira, Simonyan, Zhang, Hillier, Vijayanarasimhan, Viola, Green, Back, Natsev, et~al.]{kay2017kinetics}
Will Kay, Joao Carreira, Karen Simonyan, Brian Zhang, Chloe Hillier, Sudheendra Vijayanarasimhan, Fabio Viola, Tim Green, Trevor Back, Paul Natsev, et~al.
\newblock The kinetics human action video dataset.
\newblock \emph{arXiv preprint arXiv:1705.06950}, 2017.

\bibitem[Larsen et~al.(2016)Larsen, Sønderby, Larochelle, and Winther]{pmlr-v48-larsen16}
Anders Boesen~Lindbo Larsen, Søren~Kaae Sønderby, Hugo Larochelle, and Ole Winther.
\newblock Autoencoding beyond pixels using a learned similarity metric.
\newblock In Maria~Florina Balcan and Kilian~Q. Weinberger (eds.), \emph{Proceedings of The 33rd International Conference on Machine Learning}, volume~48 of \emph{Proceedings of Machine Learning Research}, pp.\  1558--1566, New York, New York, USA, 20--22 Jun 2016. PMLR.

\bibitem[Lee et~al.(2019)Lee, Lee, Kim, Kosiorek, Choi, and Teh]{pmlr-v97-lee19d}
Juho Lee, Yoonho Lee, Jungtaek Kim, Adam Kosiorek, Seungjin Choi, and Yee~Whye Teh.
\newblock Set transformer: A framework for attention-based permutation-invariant neural networks.
\newblock In \emph{Proceedings of the 36th International Conference on Machine Learning}, volume~97 of \emph{Proceedings of Machine Learning Research}, pp.\  3744--3753. PMLR, 09--15 Jun 2019.

\bibitem[Lin et~al.(2013)Lin, Chen, and Yan]{lin2013network}
Min Lin, Qiang Chen, and Shuicheng Yan.
\newblock Network in network.
\newblock \emph{arXiv preprint arXiv:1312.4400}, 2013.

\bibitem[Loshchilov \& Hutter(2017)Loshchilov and Hutter]{loshchilov2017decoupled}
Ilya Loshchilov and Frank Hutter.
\newblock Decoupled weight decay regularization.
\newblock \emph{arXiv preprint arXiv:1711.05101}, 2017.

\bibitem[Miech et~al.(2019)Miech, Zhukov, Alayrac, Tapaswi, Laptev, and Sivic]{miech2019howto100m}
Antoine Miech, Dimitri Zhukov, Jean-Baptiste Alayrac, Makarand Tapaswi, Ivan Laptev, and Josef Sivic.
\newblock Howto100m: Learning a text-video embedding by watching hundred million narrated video clips.
\newblock In \emph{Proceedings of the IEEE/CVF international conference on computer vision}, pp.\  2630--2640, 2019.

\bibitem[Olsson et~al.(2022)Olsson, Elhage, Nanda, Joseph, DasSarma, Henighan, Mann, Askell, Bai, Chen, et~al.]{olsson2022context}
Catherine Olsson, Nelson Elhage, Neel Nanda, Nicholas Joseph, Nova DasSarma, Tom Henighan, Ben Mann, Amanda Askell, Yuntao Bai, Anna Chen, et~al.
\newblock In-context learning and induction heads.
\newblock \emph{arXiv preprint arXiv:2209.11895}, 2022.

\bibitem[Oquab et~al.(2023)Oquab, Darcet, Moutakanni, Vo, Szafraniec, Khalidov, Fernandez, Haziza, Massa, El-Nouby, et~al.]{oquab2023dinov2}
Maxime Oquab, Timoth{\'e}e Darcet, Th{\'e}o Moutakanni, Huy Vo, Marc Szafraniec, Vasil Khalidov, Pierre Fernandez, Daniel Haziza, Francisco Massa, Alaaeldin El-Nouby, et~al.
\newblock Dinov2: Learning robust visual features without supervision.
\newblock \emph{arXiv preprint arXiv:2304.07193}, 2023.

\bibitem[Parmar et~al.(2018)Parmar, Vaswani, Uszkoreit, Kaiser, Shazeer, Ku, and Tran]{parmar2018image}
Niki Parmar, Ashish Vaswani, Jakob Uszkoreit, Lukasz Kaiser, Noam Shazeer, Alexander Ku, and Dustin Tran.
\newblock Image transformer.
\newblock In \emph{International conference on machine learning}, pp.\  4055--4064. PMLR, 2018.

\bibitem[Pont-Tuset et~al.(2017)Pont-Tuset, Perazzi, Caelles, Arbel{\'a}ez, Sorkine-Hornung, and Van~Gool]{pont20172017}
Jordi Pont-Tuset, Federico Perazzi, Sergi Caelles, Pablo Arbel{\'a}ez, Alex Sorkine-Hornung, and Luc Van~Gool.
\newblock The 2017 davis challenge on video object segmentation.
\newblock \emph{arXiv preprint arXiv:1704.00675}, 2017.

\bibitem[Radford et~al.(2018)Radford, Narasimhan, Salimans, and Sutskever]{Radford2018}
Alec Radford, Karthik Narasimhan, Tim Salimans, and Ilya Sutskever.
\newblock Improving language understanding by generative pre-training.
\newblock 2018.

\bibitem[Radford et~al.(2019)Radford, Wu, Child, Luan, Amodei, Sutskever, et~al.]{Radford2019}
Alec Radford, Jeffrey Wu, Rewon Child, David Luan, Dario Amodei, Ilya Sutskever, et~al.
\newblock Language models are unsupervised multitask learners.
\newblock 2019.

\bibitem[Radosavovic et~al.(2022)Radosavovic, Xiao, James, Abbeel, Malik, and Darrell]{radosavovic2023real}
Ilija Radosavovic, Tete Xiao, Stephen James, Pieter Abbeel, Jitendra Malik, and Trevor Darrell.
\newblock Real-world robot learning with masked visual pre-training.
\newblock In \emph{Conference on Robot Learning}, 2022.

\bibitem[Radosavovic et~al.(2023)Radosavovic, Shi, Fu, Goldberg, Darrell, and Malik]{radosavovic2023robot}
Ilija Radosavovic, Baifeng Shi, Letian Fu, Ken Goldberg, Trevor Darrell, and Jitendra Malik.
\newblock Robot learning with sensorimotor pre-training.
\newblock In \emph{Conference on Robot Learning}, 2023.

\bibitem[Ragusa et~al.(2023)Ragusa, Farinella, and Furnari]{ragusa2023stillfast}
Francesco Ragusa, Giovanni~Maria Farinella, and Antonino Furnari.
\newblock Stillfast: An end-to-end approach for short-term object interaction anticipation.
\newblock In \emph{Proceedings of the IEEE/CVF Conference on Computer Vision and Pattern Recognition}, pp.\  3635--3644, 2023.

\bibitem[Ramesh et~al.(2021)Ramesh, Pavlov, Goh, Gray, Voss, Radford, Chen, and Sutskever]{ramesh2021zero}
Aditya Ramesh, Mikhail Pavlov, Gabriel Goh, Scott Gray, Chelsea Voss, Alec Radford, Mark Chen, and Ilya Sutskever.
\newblock Zero-shot text-to-image generation.
\newblock In \emph{International Conference on Machine Learning}, pp.\  8821--8831. PMLR, 2021.

\bibitem[Ranzato et~al.(2014)Ranzato, Szlam, Bruna, Mathieu, Collobert, and Chopra]{ranzato2014video}
MarcAurelio Ranzato, Arthur Szlam, Joan Bruna, Michael Mathieu, Ronan Collobert, and Sumit Chopra.
\newblock Video (language) modeling: a baseline for generative models of natural videos.
\newblock \emph{arXiv preprint arXiv:1412.6604}, 2014.

\bibitem[Rombach et~al.(2022)Rombach, Blattmann, Lorenz, Esser, and Ommer]{rombach2022high}
Robin Rombach, Andreas Blattmann, Dominik Lorenz, Patrick Esser, and Bj{\"o}rn Ommer.
\newblock High-resolution image synthesis with latent diffusion models.
\newblock In \emph{Proceedings of the IEEE/CVF conference on computer vision and pattern recognition}, pp.\  10684--10695, 2022.

\bibitem[Russakovsky et~al.(2015)Russakovsky, Deng, Su, Krause, Satheesh, Ma, Huang, Karpathy, Khosla, Bernstein, et~al.]{russakovsky2015imagenet}
Olga Russakovsky, Jia Deng, Hao Su, Jonathan Krause, Sanjeev Satheesh, Sean Ma, Zhiheng Huang, Andrej Karpathy, Aditya Khosla, Michael Bernstein, et~al.
\newblock Imagenet large scale visual recognition challenge.
\newblock \emph{International journal of computer vision}, 115:\penalty0 211--252, 2015.

\bibitem[Ryali et~al.(2023)Ryali, Hu, Bolya, Wei, Fan, Huang, Aggarwal, Chowdhury, Poursaeed, Hoffman, et~al.]{ryali2023hiera}
Chaitanya Ryali, Yuan-Ting Hu, Daniel Bolya, Chen Wei, Haoqi Fan, Po-Yao Huang, Vaibhav Aggarwal, Arkabandhu Chowdhury, Omid Poursaeed, Judy Hoffman, et~al.
\newblock Hiera: A hierarchical vision transformer without the bells-and-whistles.
\newblock \emph{arXiv preprint arXiv:2306.00989}, 2023.

\bibitem[Shan et~al.(2020)Shan, Geng, Shu, and Fouhey]{shan2020understanding}
Dandan Shan, Jiaqi Geng, Michelle Shu, and David~F Fouhey.
\newblock Understanding human hands in contact at internet scale.
\newblock In \emph{Proceedings of the IEEE/CVF conference on computer vision and pattern recognition}, pp.\  9869--9878, 2020.

\bibitem[Shannon(1951)]{shannon1951prediction}
Claude~E Shannon.
\newblock Prediction and entropy of printed english.
\newblock \emph{Bell system technical journal}, 1951.

\bibitem[Shazeer(2020)]{shazeer2020glu}
Noam Shazeer.
\newblock Glu variants improve transformer.
\newblock \emph{arXiv preprint arXiv:2002.05202}, 2020.

\bibitem[Simonyan \& Zisserman(2014)Simonyan and Zisserman]{simonyan2014very}
Karen Simonyan and Andrew Zisserman.
\newblock Very deep convolutional networks for large-scale image recognition.
\newblock \emph{arXiv preprint arXiv:1409.1556}, 2014.

\bibitem[Su et~al.(2024)Su, Ahmed, Lu, Pan, Bo, and Liu]{su2024roformer}
Jianlin Su, Murtadha Ahmed, Yu~Lu, Shengfeng Pan, Wen Bo, and Yunfeng Liu.
\newblock Roformer: Enhanced transformer with rotary position embedding.
\newblock \emph{Neurocomputing}, 568:\penalty0 127063, 2024.

\bibitem[Touvron et~al.(2023)Touvron, Lavril, Izacard, Martinet, Lachaux, Lacroix, Rozi{\`e}re, Goyal, Hambro, Azhar, et~al.]{touvron2023llama}
Hugo Touvron, Thibaut Lavril, Gautier Izacard, Xavier Martinet, Marie-Anne Lachaux, Timoth{\'e}e Lacroix, Baptiste Rozi{\`e}re, Naman Goyal, Eric Hambro, Faisal Azhar, et~al.
\newblock Llama: Open and efficient foundation language models.
\newblock \emph{arXiv preprint arXiv:2302.13971}, 2023.

\bibitem[Van~den Oord et~al.(2016)Van~den Oord, Kalchbrenner, Espeholt, Vinyals, Graves, et~al.]{van2016conditional}
Aaron Van~den Oord, Nal Kalchbrenner, Lasse Espeholt, Oriol Vinyals, Alex Graves, et~al.
\newblock Conditional image generation with pixelcnn decoders.
\newblock \emph{Advances in neural information processing systems}, 29, 2016.

\bibitem[Van Den~Oord et~al.(2016)Van Den~Oord, Kalchbrenner, and Kavukcuoglu]{van2016pixel}
A{\"a}ron Van Den~Oord, Nal Kalchbrenner, and Koray Kavukcuoglu.
\newblock Pixel recurrent neural networks.
\newblock In \emph{International conference on machine learning}, pp.\  1747--1756. PMLR, 2016.

\bibitem[Vaswani et~al.(2017)Vaswani, Shazeer, Parmar, Uszkoreit, Jones, Gomez, Kaiser, and Polosukhin]{vaswani2017attention}
Ashish Vaswani, Noam Shazeer, Niki Parmar, Jakob Uszkoreit, Llion Jones, Aidan~N Gomez, {\L}ukasz Kaiser, and Illia Polosukhin.
\newblock Attention is all you need.
\newblock \emph{Advances in neural information processing systems}, 30, 2017.

\bibitem[Wang et~al.(2023{\natexlab{a}})Wang, Huang, Zhao, Tong, He, Wang, Wang, and Qiao]{wang2023videomae}
Limin Wang, Bingkun Huang, Zhiyu Zhao, Zhan Tong, Yinan He, Yi~Wang, Yali Wang, and Yu~Qiao.
\newblock Videomae v2: Scaling video masked autoencoders with dual masking.
\newblock In \emph{Proceedings of the IEEE/CVF Conference on Computer Vision and Pattern Recognition}, pp.\  14549--14560, 2023{\natexlab{a}}.

\bibitem[Wang et~al.(2023{\natexlab{b}})Wang, Chen, Wu, Chen, Dai, Liu, Yuan, and Jiang]{wang2022masked}
Rui Wang, Dongdong Chen, Zuxuan Wu, Yinpeng Chen, Xiyang Dai, Mengchen Liu, Lu~Yuan, and Yu-Gang Jiang.
\newblock Masked video distillation: Rethinking masked feature modeling for self-supervised video representation learning.
\newblock In \emph{CVPR}, 2023{\natexlab{b}}.

\bibitem[Wang et~al.(2022)Wang, Li, Li, He, Huang, Zhao, Zhang, Xu, Liu, Wang, et~al.]{wang2022internvideo}
Yi~Wang, Kunchang Li, Yizhuo Li, Yinan He, Bingkun Huang, Zhiyu Zhao, Hongjie Zhang, Jilan Xu, Yi~Liu, Zun Wang, et~al.
\newblock Internvideo: General video foundation models via generative and discriminative learning.
\newblock \emph{arXiv preprint arXiv:2212.03191}, 2022.

\bibitem[Weissenborn et~al.(2019)Weissenborn, T{\"a}ckstr{\"o}m, and Uszkoreit]{weissenborn2019scaling}
Dirk Weissenborn, Oscar T{\"a}ckstr{\"o}m, and Jakob Uszkoreit.
\newblock Scaling autoregressive video models.
\newblock \emph{arXiv preprint arXiv:1906.02634}, 2019.

\bibitem[Wu et~al.(2018)Wu, Xiong, Yu, and Lin]{wu2018unsupervised}
Zhirong Wu, Yuanjun Xiong, Stella~X Yu, and Dahua Lin.
\newblock Unsupervised feature learning via non-parametric instance discrimination.
\newblock In \emph{Proceedings of the IEEE conference on computer vision and pattern recognition}, pp.\  3733--3742, 2018.

\bibitem[Xiao et~al.(2022)Xiao, Radosavovic, Darrell, and Malik]{xiao2022masked}
Tete Xiao, Ilija Radosavovic, Trevor Darrell, and Jitendra Malik.
\newblock Masked visual pre-training for motor control.
\newblock \emph{arXiv preprint arXiv:2203.06173}, 2022.

\bibitem[Yang et~al.(2022)Yang, Hu, Babuschkin, Sidor, Liu, Farhi, Ryder, Pachocki, Chen, and Gao]{yang2022tensorprogramsvtuning}
Greg Yang, Edward~J. Hu, Igor Babuschkin, Szymon Sidor, Xiaodong Liu, David Farhi, Nick Ryder, Jakub Pachocki, Weizhu Chen, and Jianfeng Gao.
\newblock Tensor programs v: Tuning large neural networks via zero-shot hyperparameter transfer, 2022.

\bibitem[Zhang \& Sennrich(2019)Zhang and Sennrich]{zhang2019root}
Biao Zhang and Rico Sennrich.
\newblock Root mean square layer normalization.
\newblock \emph{Advances in Neural Information Processing Systems}, 32, 2019.

\bibitem[Zhang(2022)]{zhang2022tfcnet}
Shiwen Zhang.
\newblock Tfcnet: Temporal fully connected networks for static unbiased temporal reasoning.
\newblock \emph{arXiv preprint arXiv:2203.05928}, 2022.

\bibitem[Zhu et~al.(2024)Zhu, Liao, Zhang, Wang, Liu, and Wang]{zhu2024vision}
Lianghui Zhu, Bencheng Liao, Qian Zhang, Xinlong Wang, Wenyu Liu, and Xinggang Wang.
\newblock Vision mamba: Efficient visual representation learning with bidirectional state space model.
\newblock \emph{arXiv preprint arXiv:2401.09417}, 2024.

\end{thebibliography}

\newpage
\appendix
\section{Appendix}

\subsection{Video Tokens for Pre-Training}
\label{app:video_training}

The next patch prediction for visual pre-training is equivalent to the next token prediction in large language models. However, most languages have a clear sequential nature, therefore there is a clear definition for the next word. This also makes the next word prediction task relatively harder, since the model requires learning to extrapolate the data. On the other hand, images and videos, especially over the spatial dimensions lack a sequential nature. We follow the previous works~\citep{chen2020generative, van2016pixel} to make the images and videos into a 1D sequence by scanning the patches in raster order. While this ordering allows for example to learn to predict the bottom half of the image from the top part of the image, in many places, the tokens can be predicted by interpolating rather than extrapolating.

\begin{figure*}[!htb]
    \centering
    \includegraphics[width=0.7\linewidth]{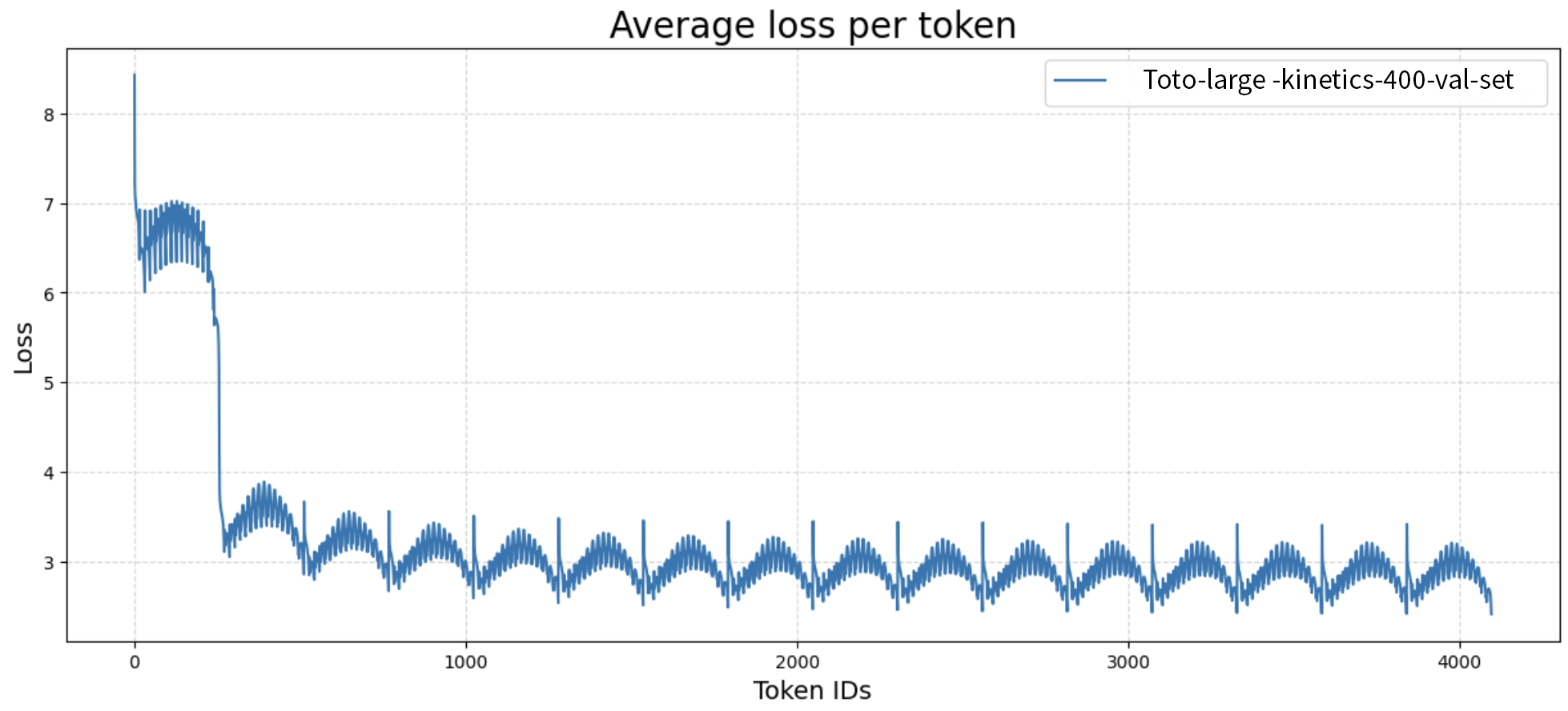}
    \caption{\textbf{Average Validation Loss Over Tokens:} We show the average loss per token for kinetics validation set. It clearly shows the redundancy in videos, as the first frame has higher prediction loss, and rest of the frames on average has lower loss than the first frame. }
    \label{fig:k400_val_set_loss}
\end{figure*}

On the time axis, yes, there is a clear sequential nature, however, video frames compared to text tokens are more redundant, making the next frame prediction task much easier. Figure~\ref{fig:k400_val_set_loss} shows average validation loss over 4096 token, in kinetics 400 dataset~\citep{kay2017kinetics}, on \approachName X-large model. This shows there is high loss of the first frame, but the subsequent frames have relatively lower loss compared to the first frame. This is because, even with reasonably lower sampling rate, frames following the first frame has some redundancy, and hinders the learning, since these tokens are relatively easy to predict. This also could be attributed by emergence of induction heads~\citep{olsson2022context}. While we focused on learning from unfiltered internet scale video with minimal inductive bias, to learn efficiently from videos, need further research in this direction.

\vspace{-0.2cm}
\subsection{Prefix attention}
During fine-tuning, we experimented with causal and full attention. On ImageNet, our base model achieved full attn: 82.6\% vs causal attn: 82.2\%. Even though our models are \textit{not pre-trained with prefix attention}, still able to utilize full attn at fine-tuning. 
This is an unrealized benefit of training with videos, (a middle token in say, 8th frame won't see the rest half of the 8th frame, but have seen all the tokens from 7th frame, which are similar because of video, hence approximating full attention at pre-training)

\vspace{-0.2cm}
\subsection{Full fine-tuning}
We fine-tuned our models on ImageNet, and performance is  close to SOTA, compared to linear probing (where we only use causal attention). But during the fine-tuning, we use full attention. 

\begin{table}[h]
\centering
\footnotesize
\begin{tabular}{cccc|c}
\toprule[0.4mm]
 \textbf{DINO} & \textbf{MoCo v3} & \textbf{BEiT} & \textbf{MAE} & \textbf{\approachName} \\ \hline
 82.8 & 83.2 & 83.2 & 83.6 & 82.6 \\ \bottomrule[0.4mm]
\end{tabular}
\caption{\textbf{Full Fine Tuning Performance:} Comparison of different methods performance on ImageNet-1K.}
\label{tab:full_finetuning}
\vspace{-0.3cm}
\end{table}

\subsection{iGPT vs \approachName X on ImagenNet}

Table~\ref{tab:imagenet} shows ImageNet evaluation performance. However, iGPT~\citep{chen2020generative} models are evaluated only using linear probing. To have a fair comparison, between iGPT and  \approachName X, we reevaluated our models using linear probing. Both models have causal attention and are trained on auto-regressive objectives. On the same model sizes, about 1 billion parameters, our achieve 66.2\% while the similar iGPT model's ImageNet performance is 65.2\%. This fair evaluation suggests the modifications made on  \approachName X have clear benefits over iGPT.

\begin{table}[!htb]
\small
\setlength{\tabcolsep}{12pt}
\centering
\begin{tabular}{llcc}
    \toprule[0.4mm]
    \textbf{Method} & \textbf{Arch} & \textbf{\#$\theta$} & \textbf{Top1} \\
    \hline
    iGPT-L~\citep{chen2020generative}  & GPT-2    & 1386  & 65.2 \\ \midrule
    \approachName X-1b               & LLaMA    & 1100  & 66.2 \\ 
    \bottomrule[0.4mm]
    \end{tabular}
    \vspace{0.2cm}
    \caption{\textbf{ImageNet Linear Probing Results:}  \approachName X performs better than similar size iGPT models.}
    \label{tab:imagenet_appendix}
\end{table}

\subsection{$\mu$-Parameterization}

To study the scaling behaviours of \approachName X using $\mu$-Parameterization~\citep{yang2022tensorprogramsvtuning}. First we train various models a1-a6 (in Table~\ref{tbl:a1_a6}), with hidden sizes (64-1536) and number of layers (12-48), increasing linearly and we used VQGAN tokenizer~\citep{esser2020taming}. Then we tune the learning rate for these models, with fixed depth using $\mu$-Parameterization~\citep{yang2022tensorprogramsvtuning}. Figure~\ref{fig:mup_loss_curves} shows optimal learning rate of $2^{-7}$ for all the model widths. Once we find the optimal learning rate, we train a1-a6 models on the mixture of image and video data, as mentioned in Table~\ref{tbl:datasets}.

\begin{table}[!htb]
\begin{center}
\footnotesize
\setlength{\tabcolsep}{4pt}
\renewcommand{\arraystretch}{2} 
\begin{minipage}{0.48\textwidth}
\centering
\begin{tabular}{c c c c c} 
\toprule[0.4mm]
\textbf{Model} & \textbf{Params} & \textbf{Dimension} & \textbf{Heads} & \textbf{Layers} \\ \midrule
a1  & 14.8M & 256  & 16 & 12 \\
a2 & 77.2M & 512 & 16 & 16 \\
a3    & 215M & 768 & 16 & 20 \\
a4    & 458M & 1024 & 16 & 24 \\
a5    & 1.2B & 1536 & 16 & 28 \\
a6    & 1.9B & 1792 & 16 & 32 \\ 
\bottomrule[0.4mm]
\end{tabular}
\caption{\textbf{\approachName X Varients:} We scale \approachName X models by increasing hidden dimension and number of layers linearly while keeping number of heads constant following \citep{yang2022tensorprogramsvtuning, touvron2023llama}.}
\label{tbl:a1_a6}
\end{minipage}
\hfill
\begin{minipage}{0.48\textwidth}
\begin{tabular}{c c c c c} 
\centering
\includegraphics[width=0.99\linewidth]{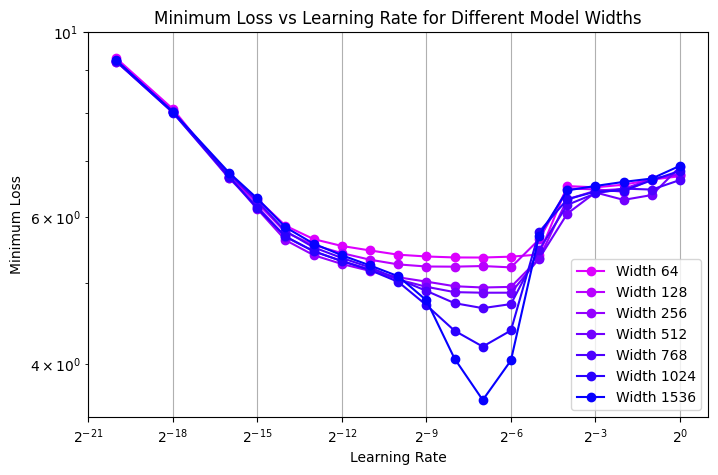}
\small
\end{tabular}
\vspace{-0.2cm}
\captionof{figure}{\textbf{$\mu$-Parameterization Learning Rate:} We show that $\mu$-Parameterization~\citep{yang2022tensorprogramsvtuning}, we can train all width \approachName X models, with an single optimal learning rate of $2^{-7}$.}
\label{fig:mup_loss_curves}
\end{minipage}
\end{center}
\end{table}

\newpage
\subsection{n-gram distribution}
\label{app:ngram}

\begin{figure*}[!htb]
    \centering
    \includegraphics[width=0.94\linewidth]{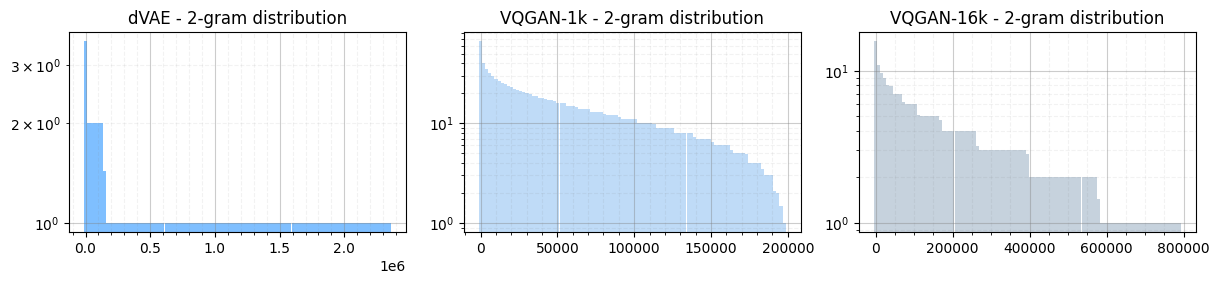}
    \caption{\textbf{2-gram Distribution of Various Tokens:} We compute the 2-gram distribution on 10000 images from the ImageNet validation set. Compared to VQGAN 1k and 16k vocabulary tokenizers, the dVAE tokenizer has a larger set of token combinations. }
    \label{fig:2-gram}
\end{figure*}

\begin{figure*}[!htb]
    \centering
    \includegraphics[width=0.94\linewidth]{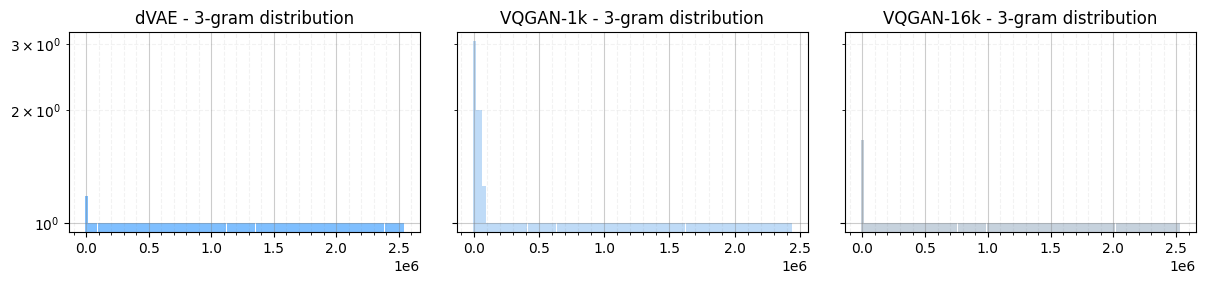}
    \caption{\textbf{3-gram Distribution of Various Tokens:} We compute the 3-gram distribution on 10000 images from the ImageNet validation set. All the tokenizers has similar almost flat distribution when it comes to 3-gram tokens.}
    \label{fig:3-gram}
\end{figure*}

In this section, we compare the 2-gram and 3-gram distribution of dVAE~\citep{ramesh2021zero}, VQGAN~\citep{esser2020taming} image tokeizers. We compute 2-gram and 3-gram distributions on the discrete tokens of 10000 ImageNet validation images. Figure~\ref{fig:2-gram} and Figure~\ref{fig:3-gram} show the distributions of these tokenizers respectively. On 2-gram distribution, dVAE~\citep{ramesh2021zero} has more discrete combination of tokens compared to both VQGAN-1K and VQGAN-16k tokenizers. 

\subsection{Attention probing variants on K400}

We also evaluate our models and baselines on the Kinetics 400 dataset using a variant of attention probing. In the main paper, we use attention probing, with only learning $W_k, W_v$ matrices, and a single learnable query vector. We also test with cross attention with MLP layers as the attention classifier, to give more capacity to the learnable head. Table~\ref{tab:k400_2}
show the performance on the attention classifier with an additional MLP head. This helps to improve performance across over all models.

\begin{table}[!htb]
\setlength{\tabcolsep}{12pt}
\small
  \centering
    \begin{tabular}{llcc}
    \toprule[0.4mm]
    \textbf{Method} & \textbf{Arch} & \textbf{Top1}  \\
    \hline
    Hiera~\citep{ryali2023hiera}                  & Hiera-L/14 & 74.2  \\ 
    Hiera~\citep{ryali2023hiera}                  & Hiera-H/14 & 75.2  \\ 
    VideoMAE~\citep{wang2023videomae}             & ViT-B/14   & 65.4  \\  
    VideoMAE~\citep{wang2023videomae}             & ViT-L/14   & 74.8  \\ \hline 
    \approachName X-base                         & LLaMA      & 61.2  \\ 
    \approachName X-large                        & LLaMA      & 65.8  \\  
    \approachName X-1b                           & LLaMA      & 74.8  \\ 
    \bottomrule[0.4mm]
    \end{tabular}
    \vspace{0.2cm}
    \caption{\textbf{K400 Results:} We evaluate our models using cross attention and MLP layer as the classification head. Overall using a high-capacity head improves the performance across all models.  }
    \label{tab:k400_2}
\end{table}

\newpage
\subsection{Generation samples}


\noindent {\bf long video generation:} we can generate up to 64 frames, first raw: periodic motion, second raw: object permanence (light stand).
\begin{figure}[!htb]
    \centering
    \includegraphics[width=0.9\linewidth]{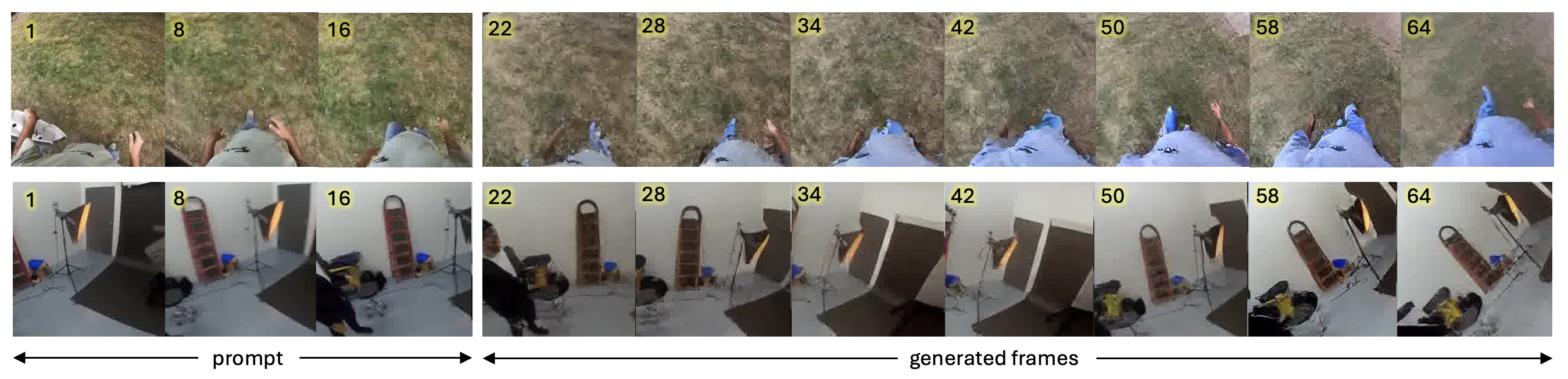}
    \label{fig:enter-label}
\end{figure}

\noindent {\bf prompting (pre-trained model):} shows 3D rotation
\begin{figure}[!htb]
    \centering
    \includegraphics[width=0.85\linewidth]{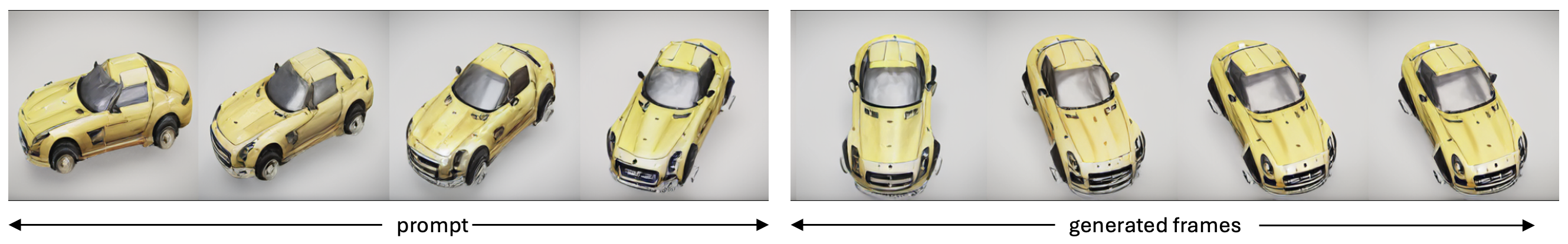}
    \label{fig:enter-label}
\end{figure}
\vspace{-0.2cm}

\noindent {\bf prompting (finetuned model):} A small 1000-step fine-tuning leads to a promptable model for various vision tasks.
\begin{figure}[!htb]
    \centering
    \includegraphics[width=0.9\linewidth]{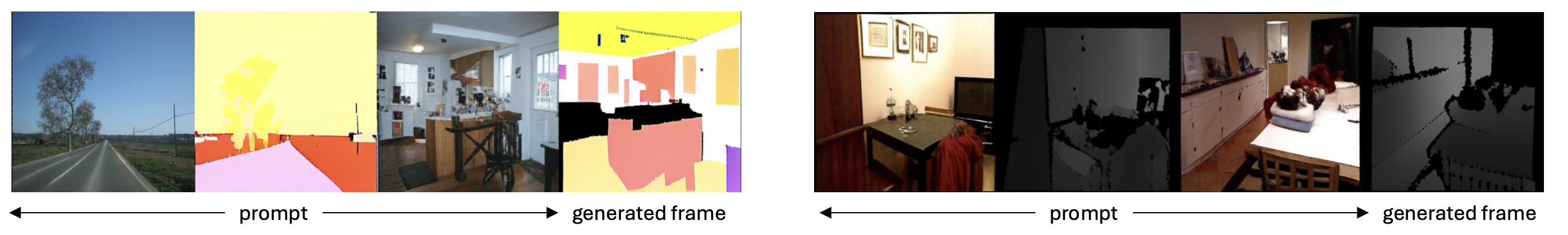}
    \label{fig:enter-label}
\end{figure}

\subsection{Additional Layer-wise Probing Results}
\label{app:more_probing}

\begin{figure}[!htb]
    \centering
    \includegraphics[width=0.4\linewidth]{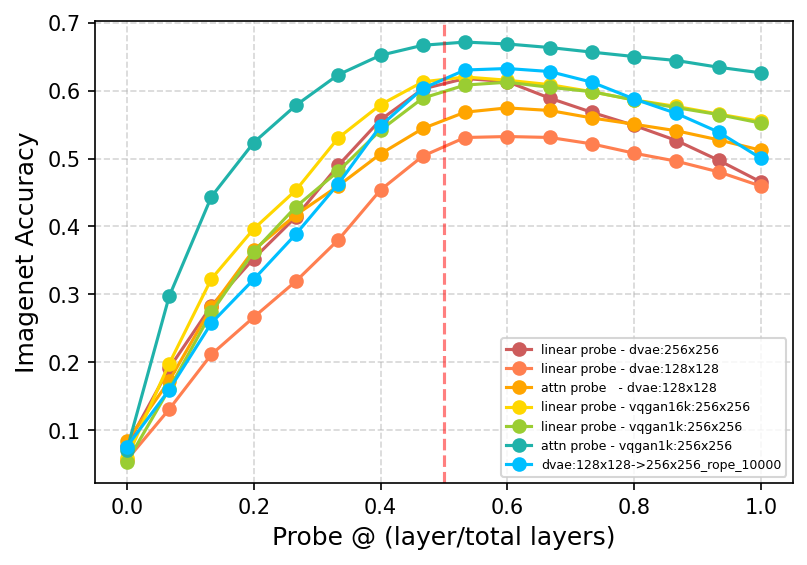}
    \caption{\textbf{Training Loss Curves:} We show the training loss curves for multiple variants of our models.}
    \label{fig:loss_curves}
\end{figure}

We probe the multiple variants of our models at each layer for the best ImageNet performance. First, we test the models on linear probing, on both sizes of 128 and 256 resolution. Figure~\ref{fig:loss_curves} presents the probing curves of the models trained with attention probing at 128 resolution. Across all models, the performance has a similar behavior to the pre-trained models, with peak performance around the middle of the depth of the model.

\end{document}